\documentclass[]{ailab}
\microtypesetup{expansion=false}
\newcommand{\our}{SAS\xspace}

\usepackage{wrapfig}
\usepackage{tabularx}
\usepackage{textcomp}
\usepackage{stfloats}
\usepackage{url}
\usepackage{verbatim}
\usepackage{titlesec}
\usepackage{tocloft}
\usepackage{adjustbox}
\usepackage{multirow}
\usepackage{pifont}
\usepackage{tikz}
\usepackage{comment}
\usepackage{amsmath,amssymb} %
\usepackage{colortbl}  %
\usepackage{color}
\usepackage{booktabs} 
\usepackage{natbib} 
\setcitestyle{square,comma,numbers,sort&compress}
\usepackage{graphicx}    %
\usepackage{subcaption} 
\usepackage{multirow} %
\usepackage{tcolorbox}
\usepackage{booktabs} %
\usepackage{subcaption} %
\usepackage[dvipsnames]{xcolor}
\usepackage{graphicx}
\usepackage{amsmath, amssymb}
\RequirePackage{xspace}
\makeatletter
\DeclareRobustCommand\onedot{\futurelet\@let@token\@onedot}
\def\@onedot{\ifx\@let@token.\else.\null\fi\xspace}

\usepackage{lineno}

\definecolor{darkblue}{rgb}{0, 0, 0.5}
\hypersetup{colorlinks=true, citecolor=darkblue, linkcolor=darkblue, urlcolor=darkblue}

\usepackage{wrapfig}  %
\usepackage{lipsum}  

\newcommand{\xmark}{\ding{55}}

\newtcbtheorem[number within=section]{exmp}{Prompts}%
{breakable,colback=white!5!white,colframe=black!95!,fonttitle=\bfseries, left=.02in, right=.02in,bottom=.02in, top=.02in}{exmp}

\newtcbtheorem[number within=section]{case_data}{Examples}%
{breakable,
 colback=white!5!white,
 colframe={rgb,255:red,78; green,136; blue,114}, %
 fonttitle=\bfseries,
 left=.02in,
 right=.02in,
 bottom=.02in,
 top=.02in}{case_data}

\definecolor{mygreen}{HTML}{85A490}
\definecolor{myred}{HTML}{C75841}

\definecolor{cotred}{RGB}{200, 30, 30}
\definecolor{ForestGreen}{RGB}{34, 139, 34}

\usepackage[normalem]{ulem}

\def\eg{\emph{e.g}\onedot} 

\def\ie{\emph{i.e}\onedot}

\makeatother

\definecolor{adptorange}{RGB}{248, 205, 172}
\definecolor{cmpblue}{RGB}{189, 215, 238}
\definecolor{cmpblue}{RGB}{189, 215, 238}

\definecolor{our_red}{RGB}{232,157,160}
\definecolor{our_blue}{RGB}{136,206,230}
\definecolor{our_orange}{RGB}{246,200,168}
\definecolor{our_green}{RGB}{178,211,164}

\definecolor{attn_code0}{RGB}{247,215,200}
\definecolor{attn_code1}{RGB}{238,169,139}
\definecolor{mlp_code0}{RGB}{204,201,221}
\definecolor{mlp_code1}{RGB}{102,95,153}

\definecolor{token_blue}{RGB}{84, 120, 140}

\usepackage{amsmath,amsfonts,bm}

\def\eqref#1{equation~\ref{#1}}

\def\1{\bm{1}}

\DeclareMathAlphabet{\mathsfit}{\encodingdefault}{\sfdefault}{m}{sl}
\SetMathAlphabet{\mathsfit}{bold}{\encodingdefault}{\sfdefault}{bx}{n}

\usepackage{amsmath,amsfonts,bm}

\def\eqref#1{equation~\ref{#1}}

\def\1{\bm{1}}

\DeclareMathAlphabet{\mathsfit}{\encodingdefault}{\sfdefault}{m}{sl}
\SetMathAlphabet{\mathsfit}{bold}{\encodingdefault}{\sfdefault}{bx}{n}

\usepackage{multirow}
\usepackage{diagbox}
\usepackage{makecell}
\usepackage{tabularx}
\usepackage{graphicx}

\usepackage{array}
\usepackage{rotating}

\definecolor{aliceblue}{rgb}{0.94, 0.97, 1.0}
\definecolor{citecolor}{HTML}{0071BC}
\definecolor{linkcolor}{HTML}{ED1C24}
\definecolor{darkgreen}{HTML}{539165}

\makeatletter
\newcommand{\thickhline}{%
 \noalign {\ifnum 0=`}\fi \hrule height 1pt
 \futurelet \reserved@a \@xhline
}
\makeatother

\usepackage{pifont}

\usepackage{pifont}       %
\usepackage{bbding}       %
\usepackage{fontawesome}
\usepackage{xspace}
\usepackage{float}
\usepackage{enumitem}

\newlength\savewidth

\newcolumntype{x}[1]{>{\centering\arraybackslash}p{#1pt}}
\newcolumntype{y}[1]{>{\raggedright\arraybackslash}p{#1pt}}
\newcolumntype{z}[1]{>{\raggedleft\arraybackslash}p{#1pt}}

\renewcommand{\paragraph}[1]{\vspace{1mm}\noindent\textbf{#1}}
\usepackage{colortbl}
\usepackage{xcolor}
\usepackage{wrapfig}

\renewcommand{\paragraph}[1]{\vspace{1.25mm}\noindent\textbf{#1}}

\usepackage{algorithm}
\usepackage{listings}

\definecolor{codeblue}{rgb}{0.25, 0.5, 0.5}
\definecolor{codekw}{rgb}{0.35, 0.35, 0.75}
\lstdefinestyle{Pytorch}{
    language = Python,
    backgroundcolor = \color{white},
    basicstyle = \fontsize{9pt}{8pt}\selectfont\ttfamily\bfseries,
    columns = fullflexible,
    aboveskip=1pt,
    belowskip=1pt,
    breaklines = true,
    captionpos = b,
    commentstyle = \color{codeblue},
    keywordstyle = \color{codekw},
}

\definecolor{colSubject}{HTML}{D32F2F}   %
\definecolor{colAction}{HTML}{F57C00}    %
\definecolor{colDetail}{HTML}{388E3C}    %
\definecolor{colSpatial}{HTML}{1976D2}   %
\definecolor{colMood}{HTML}{7B1FA2}      %
\definecolor{colKnow}{HTML}{AFB42B}      %

\usepackage{xcolor}

\usepackage{tikz}
\usetikzlibrary{tikzmark, calc, shadows.blur, shapes.geometric, fit, positioning}
\usepackage{xparse}
\pgfdeclarelayer{background}
\pgfdeclarelayer{floatbox}
\pgfdeclarelayer{foreground}
\pgfsetlayers{background,floatbox,main,foreground}

\newcounter{hcellcount}

\NewDocumentCommand{\hctext}{m}{\csname hctext@#1\endcsname}
\NewDocumentCommand{\sethctext}{mm}{\expandafter\gdef\csname hctext@#1\endcsname{#2}}

\definecolor{scoreRed}{RGB}{200, 0, 0}
\definecolor{grayText}{RGB}{120, 120, 120}

\definecolor{green}{HTML}{009000}
\definecolor{red}{HTML}{ea4335}

\definecolor{cvblue}{rgb}{0.15, 0.45, 0.68}

\usepackage{mathtools}
\usepackage{nicefrac}
\usepackage{microtype}
\usepackage{algpseudocode}
\usepackage{longtable}
\usepackage{ragged2e}

\definecolor{codebg}{HTML}{F7F8FA}
\definecolor{codeframe}{HTML}{D0D7DE}

\usepackage[most]{tcolorbox}
\tcbuselibrary{listings,skins,breakable}

\usepackage[scaled=0.85]{beramono}

\definecolor{codebg}{HTML}{FAFBFC}      %
\definecolor{codeframe}{HTML}{E1E4E8}   %
\definecolor{codekw}{HTML}{0033B3}      %
\definecolor{codecmt}{HTML}{8A8F98}     %
\definecolor{codestr}{HTML}{067D17}     %
\definecolor{codenum}{HTML}{B0B6BC}     %

\newtcblisting{pycodebox}{
  listing engine=listings,
  listing options={
    language=Python,
    basicstyle=\scriptsize\ttfamily,
    keywordstyle=\color{codekw},
    commentstyle=\color{codecmt}\itshape,
    stringstyle=\color{codestr},
    numbers=left,
    numberstyle=\tiny\color{codenum},
    numbersep=8pt,
    xleftmargin=16pt,
    breaklines=true,
    breakatwhitespace=false,
    showstringspaces=false,
    tabsize=4,
    columns=fullflexible,
    keepspaces=true,
    upquote=true,
    aboveskip=2pt,
    belowskip=2pt,
  },
  listing only,
  enhanced,
  colback=codebg,
  colframe=codeframe,
  boxrule=0.4pt,
  arc=3pt,
  left=1mm,
  right=2mm,
  top=1.5mm,
  bottom=1.5mm,
  boxsep=0pt
}

\newcommand{\std}[1]{{\scriptsize (#1)}}
\newcommand{\valstd}[2]{#1\,\std{#2}}
\newcommand{\bvalstd}[2]{\textbf{#1}\,\std{#2}}

\providecommand{\xmark}{\ding{55}}

\title{\centering \our: Simple Attention Sparsification via End-to-End Optimization of Context Ranking}

\author[1,2,*]{Zhiwei Li}
\author[1,*,\dag]{Lei Zhu}
\author[1,3]{Hao Gu}
\author[1]{Xiang Hu}
\author[1]{Yan Wang}
\author[1]{Haitao Mi}
\author[3]{Sirui Han}
\author[1]{Leowei Liang}
\author[2,3]{Zhijiang Guo}

\affiliation[1]{Tencent HY LLM Frontier}
\affiliation[2]{Hong Kong University of Science and Technology (Guangzhou)}
\affiliation[3]{Hong~Kong University of Science and Technology}

\contribution[*]{Equal Contribution}
\contribution[\dag]{Corresponding Author}

\abstract{
    Post-training attention sparsification reduces the quadratic cumulative attention cost of pretrained Transformers by selecting a small set of context units (tokens or blocks) for each query.
    Existing trainable methods usually rely on a lightweight selector to score context units followed by a hard Top-$K$ selection, which blocks gradients from the language modeling loss. As a result, these methods commonly resort to distilling layer-wise dense attention distributions.
    While this approach encourages the selector to rank context units according to dense attention weights in the original model, such a ranking is not directly aligned with their impact on the model's final predictions under a fixed attention budget (\ie, the number of attended context units per query), which can waste the limited attention budget on less useful units.
    To address this ranking misalignment, we propose \textbf{S}imple \textbf{A}ttention \textbf{S}parsification (\our), a gated sparse attention mechanism that optimizes context ranking \textit{end-to-end} with the language modeling loss.
    The key idea is to inject the selector's continuous scores into the attention logits during training, allowing the language modeling loss to update the selector through standard backpropagation.
    We identify several choices that are crucial to make this simple design work well in practice: placing the gate inside the attention $\operatorname{softmax}$ in log form,
    using normalized $\operatorname{softmax}$ gates to calibrate historical context against the always-retained current block, and preserving continuous selector scores so the model learns relative priorities rather than only hard selections.
    To support long-sequence training, we implement a memory-efficient Triton kernel that integrates our method into FlashAttention-style computation.
    Across reasoning, long-context understanding, and agentic tasks, \our consistently outperforms trainable sparse attention baselines across attention budgets, with especially large gains under tight budgets, demonstrating more effective context ranking for downstream tasks.
}

\date{\today}
\code{\href{https://github.com/Tencent-Hunyuan/Simple-Attention-Sparsification}{https://github.com/Tencent-Hunyuan/Simple-Attention-Sparsification}}

\begin{document}
\thispagestyle{firstheader}
\maketitle

\section{Introduction}
\label{sec:intro}
Long-context inference has become a critical efficiency bottleneck for LLMs, as autoregressive generation requires dense attention over all preceding context tokens when generating each new token. This makes the cumulative attention cost grow quadratically with context length.
Since many high-performing LLMs are already deployed with dense attention, practice favors inference-efficient methods that avoid architectural changes or retraining.
This motivates post-training attention sparsification\footnote{We use ``post-training'' relative to dense pretraining: it includes methods applied after a dense model has been pretrained, such as mid-training adaptation, post-training fine-tuning, and training-free inference-time sparsification.} 
\citep{tang2024quest, liu2025deepseek, gao2024seerattention, gao2025seerattention}, which adapts dense models into sparse-attention models after pretraining.

\begin{figure}[!t]
    \centering
    \includegraphics[width=0.99\linewidth]{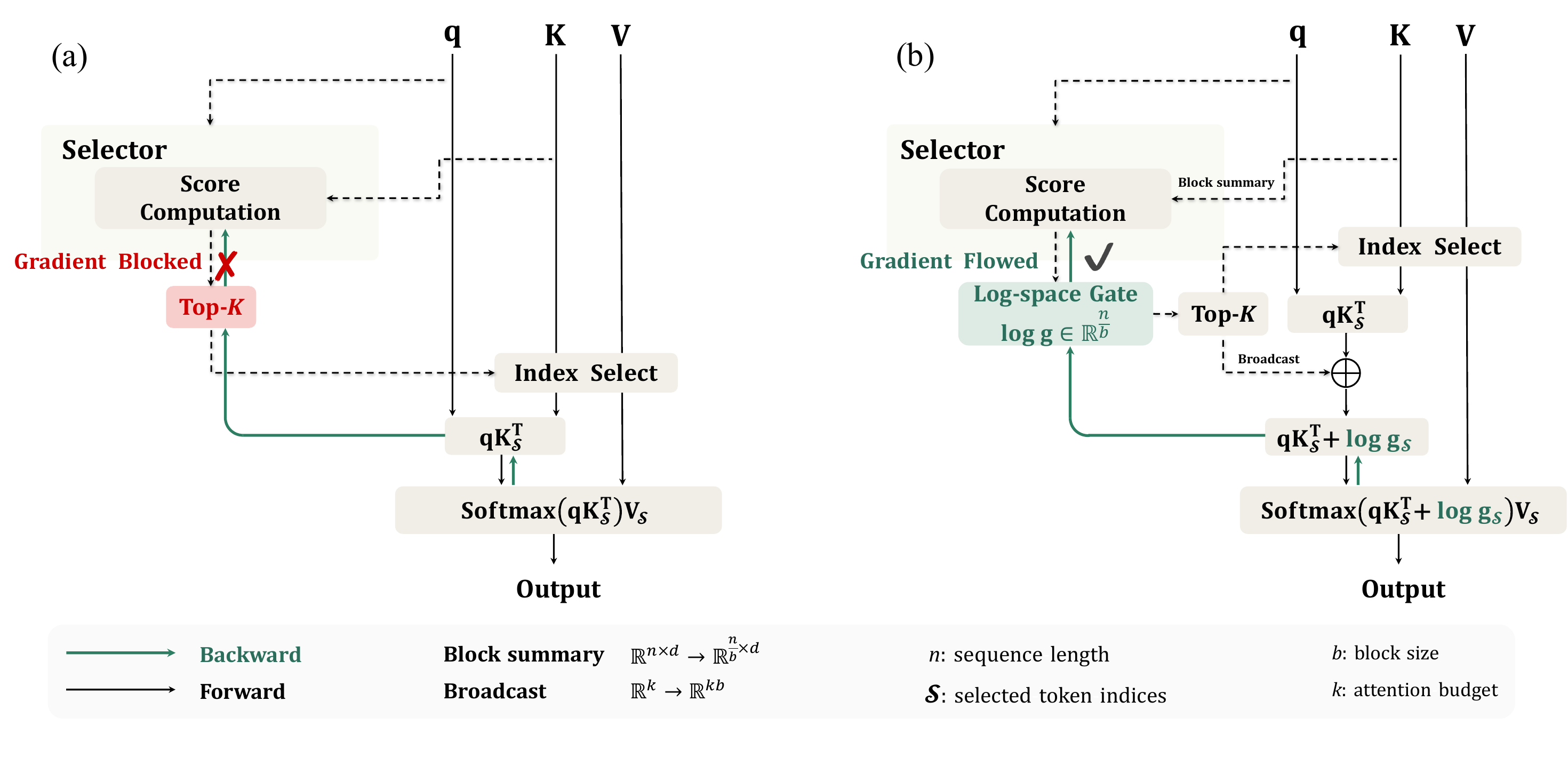}
    \caption{(a) Gradient Blockage in Discrete Selection. Sparse attention with \textbf{discrete Top-$K$ selection} prevents gradients propagation from language modeling loss to selector. Existing approaches bypass this issue using \emph{auxiliary distillation} \citep{deepseekai2026deepseekv4,gao2025seerattention} or \emph{heuristics} \citep{lumoba,zhao2025infllm}. 
    (b) Differentiable Continuous Gating. \our keeps discrete Top-$K$ selection but attaches a continuous soft gate to each selected block, enabling gradient flow from the language modeling loss to the selector.}
    \label{fig:motivation}
\end{figure}

In post-training attention sparsification for LLMs, the central question is how to select the most useful context units (tokens or blocks) for each query under a limited attention budget, where a unit can be an individual token or a coarse-grained block depending on the sparsification scheme.
Many training-free methods address this problem with heuristic or empirical rules~\citep{xiao2023streamingllm,tang2024quest,xiao2024duoattention,lin2025twilight,jiang2024minference}. 
While these training-free methods are attractive for their simplicity, their hand-crafted selection rules cannot adapt the selection strategy to the model's own prediction behavior.
More recently, trainable sparsification has emerged as a promising direction by introducing learnable selectors~\citep{gao2024seerattention,gao2025seerattention,liu2025deepseek}. A selector is a lightweight module that scores context units for each query, after which a hard Top-$K$ step chooses the attended units, enabling a higher performance ceiling than hand-crafted selection rules.
Because hard Top-$K$ selection is non-differentiable, these methods commonly train selectors through layer-wise distillation of dense attention distributions: the selector in each layer is supervised to match the attention mass produced by the original dense model.
However, this surrogate supervision can induce a ranking misalignment: it teaches the selector to rank context units according to where the original dense model attends, rather than by their impact on the model's final predictions under a limited attention budget.
This distillation strategy has two limitations. First, per-layer targets can overlook cross-layer complementarity in context usage. Second, attention matching constrains only attention weights, ignoring how attended \emph{values} (the $\mathbf{V}$ matrix in \Cref{fig:motivation}) affect the final prediction.

In this work, we explore \emph{end-to-end} optimization of context selection with the language modeling loss.
Our key idea is to make the selector's scores part of the differentiable attention computation during training, so that the standard language modeling loss can update the selector through backpropagation.
We instantiate this idea in the open-source block sparse setting of SeerAttention-R~\citep{gao2025seerattention}, where layer-wise selectors score context blocks for each query.
Instead of using layer-wise attention distillation as in SeerAttention-R, we optimize the selectors end-to-end via a simple design: during training, blockwise scores from selector are interpreted as log-space gates, and added to the $\mathbf{q}\mathbf{K}^\top$ attention logits, allowing gradients to flow from the attention computation to the selectors (\Cref{fig:motivation}b).

The effectiveness of this simple design, however, depends on four key choices in how the selector is trained from the language modeling loss.
\emph{(1)} The gate should be added inside the attention $\operatorname{softmax}$ in log form, so selector scores directly modulate attention mass rather than merely rescaling attention outputs.
\emph{(2)} The gate activation should normalize scores across context blocks (\eg, with a $\operatorname{softmax}$), so the gates encode calibrated relative importance rather than unnormalized scores.
\emph{(3)} The selected gates should stay continuous rather than being collapsed into binary Top-$K$ indicators, because the model must learn not only which context blocks are selected but also their relative priorities.
\emph{(4)} Training should update only the selected blocks rather than all context blocks, since it is much cheaper and still reaches comparable final performance.
Together, the first three choices make end-to-end gradients from the language modeling loss informative for learning which context blocks actually support the model's prediction, while the last keeps training efficient.

Besides the design choices above that make the end-to-end optimization effective, we also implement specialized Triton \citep{tillet2019triton} kernels that make long-context training practical. Since a naive implementation would materialize the full attention matrix before adding the log-space gates, which is prohibitive for long-context training, our kernel fuses the gate addition into the tile-level $\mathbf{q}\mathbf{K}^\top$ computation in FlashAttention~\citep{dao2022flashattention}. As a result, attention sparsification with \our is simplified to replacing the attention kernel with our gated-attention kernel and optimizing the standard language modeling loss, without auxiliary distillation.

Despite its simplicity, extensive experiments demonstrate that \our consistently outperforms existing post-training attention sparsification baselines across reasoning, long-context understanding, and agentic tasks. We further provide initial evidence that the same end-to-end optimization applies to continued pretraining, where the backbone and selector are trained jointly.
The improvements are most pronounced under low attention budgets, where accurate context ranking is most critical: on reasoning tasks at a 1024 token budget, \our improves over SeerAttention-R~\citep{gao2025seerattention} by 6.0--7.7 points on MATH500 and 10.6--15.5 points on GPQA-Diamond across Qwen3-4B, 8B, and 14B. The gains extend beyond reasoning: on long-context understanding (LongBench) \our leads SeerAttention-R at every budget, with the largest margins on the longest inputs (\eg, +3.2 on the 8K+ bucket of Qwen3-4B at budget 2048), and on agentic tasks it improves BFCL by up to +3.5 and remains ahead on VitaBench, nearly recovering full attention at budget 4096.
Our contributions are:
{\setlength{\leftmargini}{1.6em}
\begin{itemize}
\item We propose \our, an end-to-end post-training attention sparsification paradigm that trains context selectors with the language modeling loss, removing the need for teacher attention or auxiliary distillation. In controlled comparisons with matched backbones, selector architectures, and training data, \our learns more effective sparse routing than layer-wise attention distillation.
\item 
We identify the design choices that make the simple gated relaxation effective and practical, including log-space gate injection, normalized gate activation, preservation of fine-grained score differences, an efficient sparse training scope, and a memory-efficient kernel for long-context training.
\item 
\our consistently improves performance across reasoning (MATH500, GPQA-Diamond, AIME24, AIME25), long-context understanding (LongBench), and agentic tasks (BFCL, VitaBench), with gains of over 10\% under low attention budgets.
\end{itemize}
}

\section{Related Work}
\label{sec:related}

\paragraph{Inference time attention sparsification / KV-cache selection.}
A practical way to reduce long-context attention cost is to sparsify attention at inference time by selecting a subset of context units without updating the backbone LLM~\citep{xiao2023streamingllm, h2o, tang2024quest, jiang2024minference, lserve, liu2024retrievalattention, zhang2025spargeattn, xu2025xattention}.
Closely related are KV-cache methods that dynamically select relevant cached tokens without irreversible eviction, which is more suitable for long reasoning generation where future token importance is hard to predict~\citep{tang2024quest, pqcache, squeezedattention, liu2024retrievalattention, cai2026rkvredundancyawarekvcache, hao2025omnikv, metaclusterkv}.
These methods are attractive because they can be applied to existing dense LLMs with little or no additional training.
However, their sparse locations are usually determined by heuristic rules, post-hoc statistics, or external retrieval/indexing procedures, leaving the selection rule largely fixed rather than learned from the model's prediction objective. In contrast, our method keeps the post-training sparsification setting but learns a lightweight block selector directly from the language modeling loss.

\paragraph{Trainable sparse attention.}
Another line of work makes sparse attention patterns trainable.
Early sparse Transformers impose fixed layouts, such as local windows, global tokens, strided connections, or block-wise patterns~\citep{child2019generating, beltagy2020longformer, zaheer2020big}, but typically require training or adapting models around prescribed sparse patterns.
Recent methods use learnable routing, gating, or dense-sparse switching modules~\citep{zhu2023biformer, yuan2025native, lumoba, team2025minicpm4, gao2024seerattention, gao2025seerattention, zhao2025infllm, liu2025deepseek}.
However, hard Top-$K$ selection prevents language-modeling gradients from directly updating the selector, so existing methods often rely on dense-attention distillation~\citep{gao2024seerattention, gao2025seerattention, liu2025deepseek} or proxy selection signals~\citep{yuan2025native, lumoba}.
In contrast, we relax hard block selection into continuous log-space gates inside attention, enabling end-to-end selector optimization with the language modeling loss.

\section{Preliminaries}
\label{sec:preliminary}

This section reviews the attention computation underlying autoregressive language models and block sparse attention, which reduces long-context inference cost by restricting each query to a subset of context blocks.

\paragraph{Standard attention.}
Let $\mathbf{q} \in \mathbb{R}^{d}$ denote a query vector of the current token, and $\mathbf{K}, \mathbf{V} \in \mathbb{R}^{n \times d}$ denote the key and value matrices of the $n$ previous tokens, which are referred to as \textit{context tokens}.
Standard attention \cite{vaswani2017attention} computes the output $\mathbf{o} \in \mathbb{R}^d$ as
\begin{equation}
    \mathbf{o} = \operatorname{softmax}(\mathbf{q} \mathbf{K}^\top) \mathbf{V}.
    \label{eqn:fula}
\end{equation}
Here we omit the scaling factor for notational simplicity. During autoregressive decoding, each new query attends to all previous keys and values. Since the number of previous tokens grows linearly with the decoding step, the cumulative attention cost over a sequence of length $n$ grows as $\mathcal{O}(n^2)$.

\paragraph{Block sparse attention.}
To mitigate this quadratic cost, block sparse attention \cite{zhangsurvey,sun2025efficient} restricts the query $\mathbf{q}$ to attend only to a selected subset of context blocks.
Let $\mathcal{S}\subseteq\{1,\ldots,n\}$ denote the selected token-index set. 
The attention output is formulated as
\begin{equation}
    \mathbf{o}
    =
    \operatorname{softmax}
    \left(
        \mathbf{q}\mathbf{K}_{\mathcal{S}}^\top
    \right)
    \mathbf{V}_{\mathcal{S}}.
    \label{eqn:spa}
\end{equation}
For a single query, this reduces the computation cost from $\mathcal{O}(n)$ to $\mathcal{O}(|\mathcal{S}|)$.
Over a sequence of length $n$, the cumulative cost is therefore reduced to $\mathcal{O}(n|\mathcal{S}|)$.
To construct $\mathcal{S}$, the $n$ context positions are partitioned into $C$ contiguous index blocks $\{B_1,\ldots,B_C\}$, where each $B_m$ denotes a set of token indices and contains $b$ tokens (i.e., $C=n/b$).
A lightweight selector $\mathcal{R}_\theta(\cdot)$ computes relevance scores $\mathbf{s}\in\mathbb{R}^{C}$ over these blocks.
The selector chooses the $K$ most important blocks from all $C$ context blocks, and the query only attends to the union of these selected blocks.
Formally, the selected block indices and the corresponding token-index set are
\begin{equation}
    \mathcal{I}
    =
    \text{Top-}K(\mathbf{s}, K),
    \qquad
    \mathcal{S}
    =
    \bigcup_{m\in\mathcal{I}} B_m .
    \label{eqn:selected_context}
\end{equation}
Under plain hard Top-$K$, the selected index set is piecewise constant with respect to selector scores. Therefore, the language modeling loss provides no useful gradient through the selection path to update the selector, as illustrated in \Cref{fig:motivation}a.
To make the selector end-to-end trainable, the core problem is how to make selector scores affect attention differentiably, so that the language modeling loss can train the selector.

\section{Method}
\label{sec:method}

\begin{figure}[!t]
    \centering
    \includegraphics[width=0.93\linewidth]{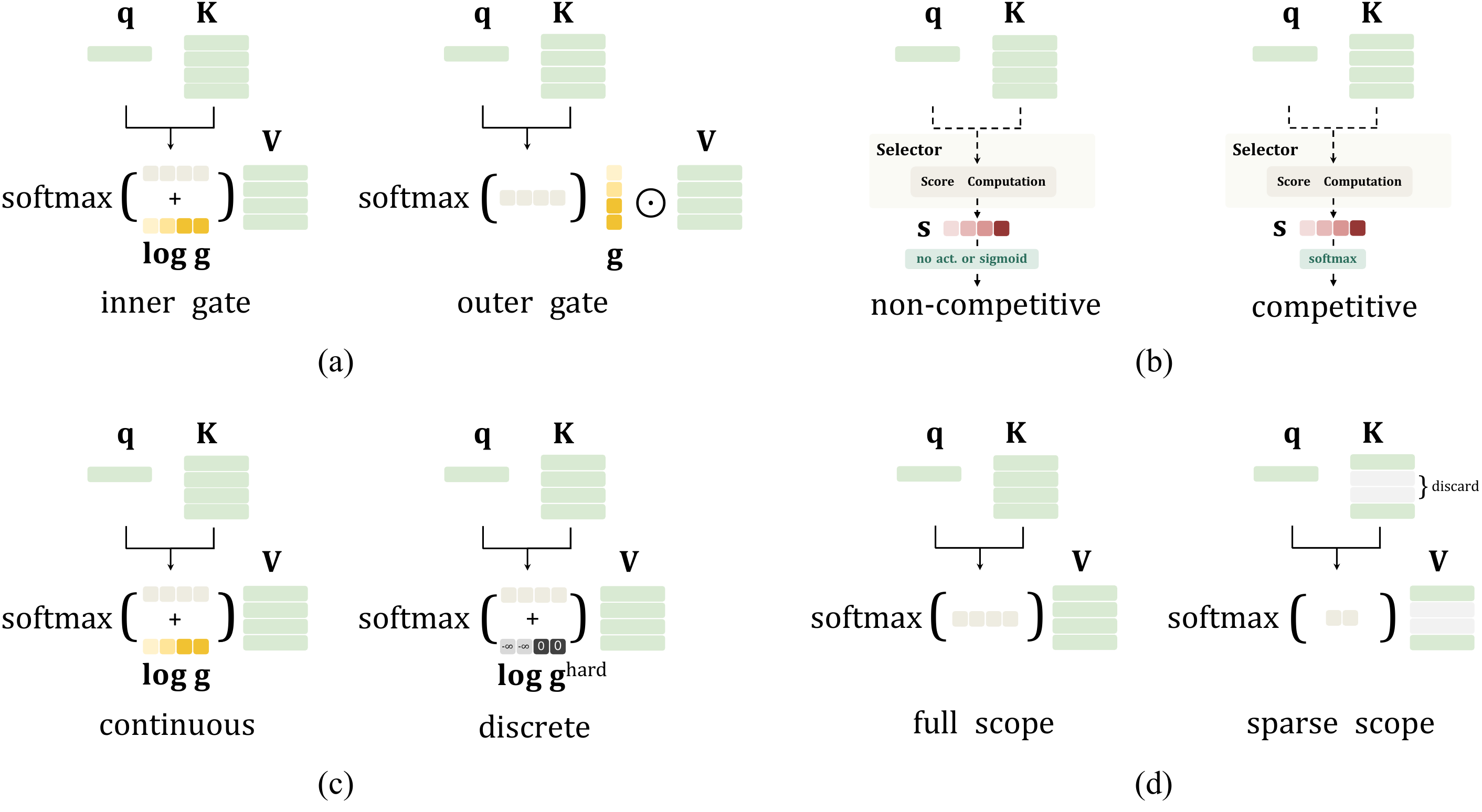}
    \caption{Components of learnable context ranking.
    (a) Gate position, whether the gates are injected into the $\operatorname{softmax}$ or applied outside the $\operatorname{softmax}$.
    (b) Gate activation, whether historical-block scores are normalized jointly with $\operatorname{softmax}$, activated independently with $\operatorname{sigmoid}$, or injected as raw logits; the current block always has a unit gate.
    (c) Ranking preservation, whether continuous scores are preserved or collapsed into binary masks.
    (d) Training scope, whether training receives the gradient from all context blocks or only the selected blocks.}
    \label{fig:abla}
\end{figure}

We present \our, a \textbf{S}imple \textbf{A}ttention \textbf{S}parsification method that trains a lightweight selector to rank context blocks from the language modeling loss.
In this section, we first reformulate sparse block selection as context ranking.
We then analyze how to make such rankings learnable from the language modeling loss, focusing on gate position, gate activation, ranking preservation, and training scope.
Finally, we introduce a specialized training kernel for the practical implementation of \our.

\subsection{From Sparse Selection to Context Ranking}
\label{subse:context_rank}

The starting point is that Top-$K$ selection is determined by the ordering of selector scores over context blocks.
Before producing a discrete selected set, the selector first defines a ranking of candidate blocks.
A good selector should assign higher ranks to blocks that are more useful for prediction.
We therefore view attention sparsification as \emph{context ranking}: during training, the selector learns a continuous ordering over blocks, and this ordering provides the basis for sparse block selection.

\paragraph{Differentiable ranking.}
To make block ranking learnable, the selector scores must affect the attention computation during training.
Following the block partition in Section~\ref{sec:preliminary}, we single out the always-retained current block as $B_0$ and treat the remaining $C$ context blocks $\mathcal{H}=\{B_1,\ldots,B_C\}$ as the candidate historical blocks that the selector ranks.
The selector produces scores $\mathbf{s}\in\mathbb{R}^{C}$ only for $\mathcal{H}$, which we convert into positive gates
\begin{equation}
    \mathbf{g} = \phi(\mathbf{s})\in\mathbb{R}_{+}^{C},
    \qquad g_0=1,
    \label{eq:general_gate}
\end{equation}
where $\phi(\cdot)$ is a differentiable activation function and $g_0=1$ leaves the current block unbiased.
This transformation allows historical-block scores to modulate attention rather than determine a discrete Top-$K$ set.

For each historical block $B_m\in\mathcal{H}$, the gate $g_m$ is broadcast to all tokens in that block, while tokens in $B_0$ retain the unit gate.
By making the attention computation depend on these broadcast gates, gradients from the language modeling loss can flow back to $\mathbf{g}$ and then to the selector.
The key design question is how to \textit{integrate gates into the attention computation so that gradients provide a reliable signal for block ranking}.

\paragraph{Components of learnable ranking.}
Merely allowing language modeling gradients to reach the selector does not guarantee effective block ranking: different gated formulations shape the resulting ranking signals differently.
As illustrated in \Cref{fig:abla}, we investigate four design elements: gate position, gate activation, ranking information preservation, and training scope.

{\setlength{\leftmargini}{2em}
\begin{itemize}
 \item  \textit{Gate position}: whether the gates are applied into the $\operatorname{softmax}$, i.e., $\mathbf{o}_{\text{inner}}=\operatorname{softmax}(\mathbf{q}\mathbf{K}^{\top}+\log\mathbf{g})\mathbf{V}$, or applied outside the $\operatorname{softmax}$, i.e., $\mathbf{o}_{\text{outer}}=\operatorname{softmax}(\mathbf{q}\mathbf{K}^{\top})(\mathbf{g}\odot\mathbf{V})$. For the inner placement, $\log(\cdot)$ ensures that the gates act as multiplicative weights on the attention probabilities after the softmax.

 \item \textit{Gate activation}: how block scores are transformed before being injected into attention. We compare normalized gates $\mathbf{g}=\operatorname{softmax}(\mathbf{s})$, independent gates $\mathbf{g}_{\sigma}=\operatorname{sigmoid}(\mathbf{s})$, and unnormalized logit injection, which adds $\mathbf{s}$ directly to attention logits. In all variants, current block remains unbiased with $g_0=1$.

 \item \textit{Ranking preservation}: whether training uses a soft gate $\mathbf{g}=\phi(\mathbf{s})$, or a hard Top-$K$ gate using STE, i.e., $\hat{\mathbf{g}}=\operatorname{stopgrad}(\mathbf{g}^{hard}-\mathbf{g})+\mathbf{g}$ with $g^{hard}_m=\mathbb{I}[m\in\text{Top-}K(\mathbf{s},K)]$.

  \item \textit{Training scope}: whether training uses the full scope in Equation~\ref{eqn:fula}, the sparse scope in Equation~\ref{eqn:spa}, or a stochastic approximation to full scope by adding noise to the scores before activation.

\end{itemize}
}

\definecolor{avg}{HTML}{D05E46}
\begin{table}[!t]
\centering
\caption{
Ablation of four gating design choices on GPQA-Diamond using Qwen3-4B with a 2048-token budget. We report avg@16 accuracy (\%),
with \textcolor{avg}{average generation length}; standard deviations (2.0--2.8) are omitted. $\mathbf{g}_{\sigma}$ denotes sigmoid-activated gates, and full$^{*}$ simulates full-scope training by adding noise before Top-$K$. Formulations show only historical-context terms; the current block always uses a unit gate.
}
\label{tab:ablation_main}

\resizebox{\textwidth}{!}{
\newcommand{\res}[2]{#1~\textcolor{avg}{\scriptsize #2}}

\begin{tabular}{l cccc | cccc}
\toprule
\textbf{Setting} &
\textbf{Gate Pos.} &
\textbf{Gate Act.} &
\textbf{Rank Pres.} &
\textbf{Train Sco.} &
\textbf{10 step} &
\textbf{100 step} &
\textbf{1000 step} &
\textbf{1 epoch}
\\
\midrule

\multicolumn{9}{c}{\textit{Baseline}}\\
\midrule

(1) $\operatorname{softmax}(\mathbf q\mathbf K^\top)\mathbf V$ & -- & -- & -- & -- &
-- & -- & -- &
\res{56.1}{8547}\\

\midrule
\multicolumn{9}{c}{\textit{I. Gate Position (inner $\operatorname{softmax}$ vs. outer $\operatorname{softmax}$)}}\\
\midrule

(2) $\operatorname{softmax}(\mathbf q\mathbf K^\top+\log\mathbf g)\mathbf V$
& inner
& $\operatorname{softmax}$
& \checkmark
& full
& \res{30.8}{25277}
& \res{\textbf{52.2}}{8696\phantom{1}}
& \res{\textbf{53.7}}{6977\phantom{1}}
& \res{{54.4}}{7109\phantom{1}}
\\

(3) $\operatorname{softmax}(\mathbf q\mathbf K^\top)(\mathbf g\odot\mathbf V)$
& outer
& $\operatorname{softmax}$
& \checkmark
& full
& \res{28.3}{25594}
& \res{38.9}{16969}
& \res{43.2}{12801}
& \res{41.6}{13807}
\\

\midrule
\multicolumn{9}{c}{\textit{II. Gate Activation ($\operatorname{softmax}$, $\operatorname{sigmoid}$, or none)}}\\
\midrule

(4) $\operatorname{softmax}(\mathbf q\mathbf K^\top+\log\mathbf g)\mathbf V$
& inner
& $\operatorname{softmax}$
& \checkmark
& full
& \res{30.8}{25277}
& \res{\textbf{52.2}}{8696\phantom{1}}
& \res{\textbf{53.7}}{6977\phantom{1}}
& \res{{54.4}}{7109\phantom{1}}
\\

(5) $\operatorname{softmax}(\mathbf q\mathbf K^\top+\log\mathbf g_{\sigma})\mathbf V$
& inner
& $\operatorname{sigmoid}$
& \checkmark
& full
& \res{19.6}{27717}
& \res{20.4}{27997}
& \res{17.6}{27989}
& \res{17.0}{28444}
\\

(6) $\operatorname{softmax}(\mathbf q\mathbf K^\top+\mathbf s)\mathbf V$
& inner
& --
& \checkmark
& full
& \res{23.4}{27540}
& \res{19.9}{27418}
& \res{18.9}{26703}
& \res{18.8}{26401}
\\

\midrule
\multicolumn{9}{c}{\textit{III. Ranking Preservation (continuous vs. discrete)}}\\
\midrule

(7) $\operatorname{softmax}(\mathbf q\mathbf K^\top+\log\mathbf g)\mathbf V$
& inner
& $\operatorname{softmax}$
& \checkmark
& full
& \res{30.8}{25277}
& \res{\textbf{52.2}}{8696\phantom{1}}
& \res{\textbf{53.7}}{6977\phantom{1}}
& \res{{54.4}}{7109\phantom{1}}
\\

(8) $\operatorname{softmax}(\mathbf q\mathbf K^\top+\log\hat{\mathbf{g}})\mathbf V$
& inner
& $\operatorname{softmax}$
& \xmark
& full
& \res{\textbf{46.5}}{9821\phantom{1}}
& \res{42.4}{8438\phantom{1}}
& \res{49.6}{7135\phantom{1}}
& \res{46.0}{7289\phantom{1}}
\\

\midrule
\multicolumn{9}{c}{\textit{IV. Training Scope (full scope vs. sparse scope)}}\\
\midrule

(9)\phantom{1} $\operatorname{softmax}(\mathbf q\mathbf K^\top+\log\mathbf g)\mathbf V$
& inner
& $\operatorname{softmax}$
& \checkmark
& full
& \res{30.8}{25277}
& \res{\textbf{52.2}}{8696\phantom{1}}
& \res{\textbf{53.7}}{6977\phantom{1}}
& \res{{54.4}}{7109\phantom{1}}
\\

(10) $\operatorname{softmax}(\mathbf q\mathbf K_{\mathcal S}^\top+\log\mathbf g_{\mathcal{S}})\mathbf V_{\mathcal S}$
& inner
& $\operatorname{softmax}$
& \checkmark
& sparse
& \res{{24.8}}{27504}
& \res{51.0}{8087\phantom{1}}
& \res{53.6}{7401\phantom{1}}
& \res{\textbf{54.8}}{7410\phantom{1}}
\\

(11) $\operatorname{softmax}(\mathbf q\mathbf K_{\tilde{\mathcal S}}^\top+\log\mathbf g_{\tilde{\mathcal{S}}})\mathbf V_{\tilde{\mathcal S}}$
& inner
& $\operatorname{softmax}$
& \checkmark
& \phantom{$^{*}$}full$^{*}$
& \res{26.3}{27350}
& \res{51.2}{7732\phantom{1}}
& \res{52.6}{7424\phantom{1}}
& \res{52.2}{7552\phantom{1}}
\\

\bottomrule
\end{tabular}
}
\end{table}

\subsection{Learning Reliable Block Ranking}
\label{subse:ranking_learn}

We next use controlled ablations to evaluate how the four design elements introduced above affect the effectiveness and stability of block-ranking optimization.

\subsubsection{Experimental setup}
\label{sec:control_setup}
We conduct controlled experiments using the AttnGate selector from SeerAttention-R \cite{gao2025seerattention}.
For our experiments, we set the block size to 64 and Top-$K$ of 32.
The experiments are performed on Qwen3-4B \cite{yang2025qwen3technicalreport} trained on 93.7K examples from OpenR1-Math-220k \cite{openr1}.
Both the training and generation lengths are set to 32{,}768 tokens.
The LLM backbone is frozen during training.
We evaluate GPQA-Diamond \cite{gpqa} with inference consistently performed under the block sparse attention formulation in Equation~\ref{eqn:spa}, and report the accuracy averaged over 16 runs and average generation length.

\subsubsection{Ablation analysis}
\label{sec:abla_analysi}
\paragraph{Overall observation.}
\Cref{tab:ablation_main} and \Cref{fig:loss_ab} compare four design dimensions for learning reliable block rankings.
First, inner $\operatorname{softmax}$ gating performs better than outer gating, showing that gates should participate in attention normalization instead of only rescaling \emph{values} ($\mathbf{V}$ matrix).
Second, normalized gating is important: $\operatorname{softmax}$ consistently outperforms independent $\operatorname{sigmoid}$ gates and unnormalized raw-logit injection because it calibrates historical context against the unit-gated current block.
Third, preserving ranking information through soft gating improves training stability, while achieving better final performance.
Unlike the first three, training scope mainly affects efficiency: sparse scope training converges more slowly at first but gradually approaches full scope, reaching comparable final performance at lower cost.

\paragraph{Gate position.}
The key  distinction is whether gates participate in attention \textit{normalization}.
Outside the $\operatorname{softmax}$, the attention probabilities are already fixed, and the gate only rescales the \emph{value} contribution of each block.
Inside the $\operatorname{softmax}$ as log-space biases, they directly change how attention mass is allocated across blocks.
Let $p_i$ denote the attention probability without gating, $\tilde p_i$ the attention probability with inner gating.
This difference is reflected in the gate gradients:
\begin{equation}
dg_m^{inner}
=
\sum_{i\in B_m}
\frac{\tilde p_i}{g_m}\,
d\mathbf{o}^{\top}
(\mathbf{v}_i-\mathbf{o}),
\qquad
dg_m^{outer}
=
\sum_{i\in B_m}
p_i\, d\mathbf{o}^{\top}\mathbf{v}_i.
\label{eq:gate_posi}
\end{equation}
Outer gating uses fixed attention probabilities, while inner gating provides a relative signal through $\mathbf{v}_i-\mathbf{o}$ for reallocating attention mass, enabling gates to learn relative importance among blocks.

\begin{figure}[!t]
    \centering
    \includegraphics[width=0.99\linewidth]{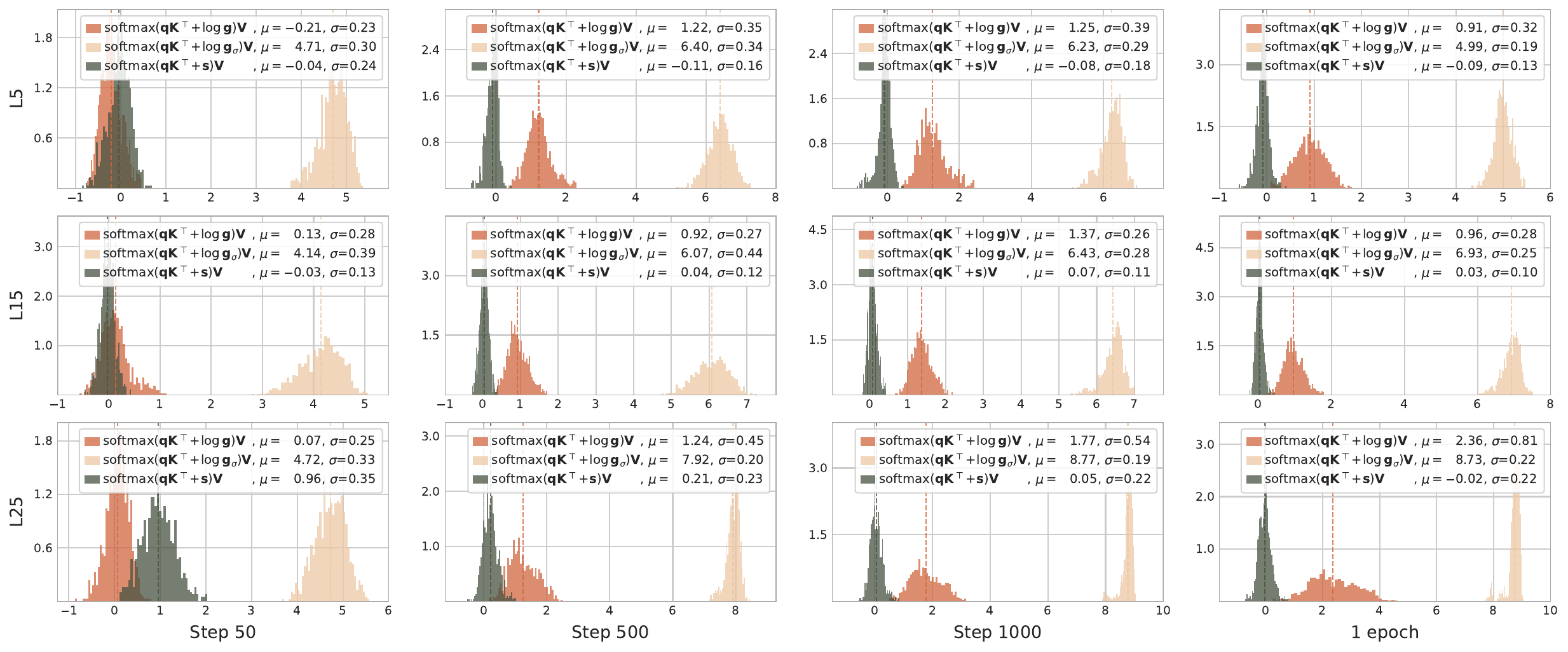}
    \caption{
        Evolution of selector logits $\mathbf{s}$ under softmax, sigmoid gating and raw-logit injection in different layers.
        Sigmoid gating saturates historical gates toward one, while raw injection collapses logits toward zero with reduced variance. Thus, both transformations approach effectively ungated attention and fail to provide a discriminative signal for learning block rankings. These observations highlight the importance of the competitive nature of softmax normalization in avoiding such trivial solutions and preserving discriminative block selection.
    }
    \label{fig:logit_dynamic}
\end{figure}

\paragraph{Activation function.}
The activation function determines how the selector calibrates historical context against the always-retained current block. For historical blocks, $\mathbf{g}=\operatorname{softmax}(\mathbf{s})$ gives
\begin{equation}
    \log g_m=s_m-\operatorname{LSE}(\mathbf{s}),
    \qquad B_m\in\mathcal{H},
    \label{eq:normalized_log_gate}
\end{equation}
while the current block has $g_0=1$ and hence zero additive bias. Although $-\operatorname{LSE}(\mathbf{s})$ is shared by all historical blocks, it does not cancel in the subsequent attention $\operatorname{softmax}$ because it is not applied to the current block. It therefore calibrates the aggregate attention mass assigned to historical context relative to the current block. This normalization also makes the gates invariant to a global shift $\mathbf{s}\mapsto\mathbf{s}+c$. By contrast, directly injecting the unnormalized logits $\mathbf{s}$ changes the historical-to-current attention balance under the same shift. Empirically, \Cref{fig:logit_dynamic} shows that $\operatorname{sigmoid}$ gates gradually saturate toward $1$, whereas unnormalized logits collapse toward $0$ with reduced variance. Both behaviors diminish the distinction between historical blocks and the unit-gated current block, weakening the learned gating signal.

\paragraph{Ranking preservation.}
The key difference is whether the forward pass preserves the continuous ranking or collapses it into a binary Top-$K$ mask.
Let $z_i=\mathbf{q}\mathbf{k}_i^\top$ denote the attention logit of token $i$, where $\mathbf{k}_i$ is its key vector, and let $b(i)$ index the block that token $i$ belongs to.
Soft gating keeps every block under a single normalizer that sums over all tokens, so each attention weight stays bounded.
Hard gating instead relies on STE for a differentiable backward: its forward normalizer sums only over the selected set $\mathcal{S}$, and a dropped block $m\notin\mathcal{I}$ is ranked by the attention mass its tokens would receive under this restricted normalizer, from which its own tokens are absent:
\begin{equation}
    \tilde p_i^{soft}
    =
    \frac{g_{b(i)}\exp(z_i)}{\sum_j g_{b(j)}\exp(z_j)}
    \le 1,
    \qquad
    \tilde p_i^{hard}
    =
    \frac{\exp(z_i)}{\sum_{j\in\mathcal{S}}\exp(z_j)}
    \le
    \exp\!\left(z_i-\max_{j\in\mathcal{S}}z_j\right).
    \label{eq:rank_unbounded}
\end{equation}
The soft weight $\tilde p_i^{soft}$ is always bounded, while the hard weight $\tilde p_i^{hard}$ has no upper bound and grows exponentially once a dropped block outscores the selected set. 
Since the gate gradient satisfies $dg_m\propto p_i$ (Equation~\ref{eq:gate_posi}), these unbounded weights are directly translated into large gradients. 
Early in training, the randomly initialized selector frequently leaves high-score blocks outside Top-$K$, causing the contributions from dropped blocks, tokens, and heads to accumulate and producing gradient norms that are orders of magnitude larger than those of soft gating.

\begin{figure}[!t]
    \centering
    \includegraphics[width=0.98\linewidth]{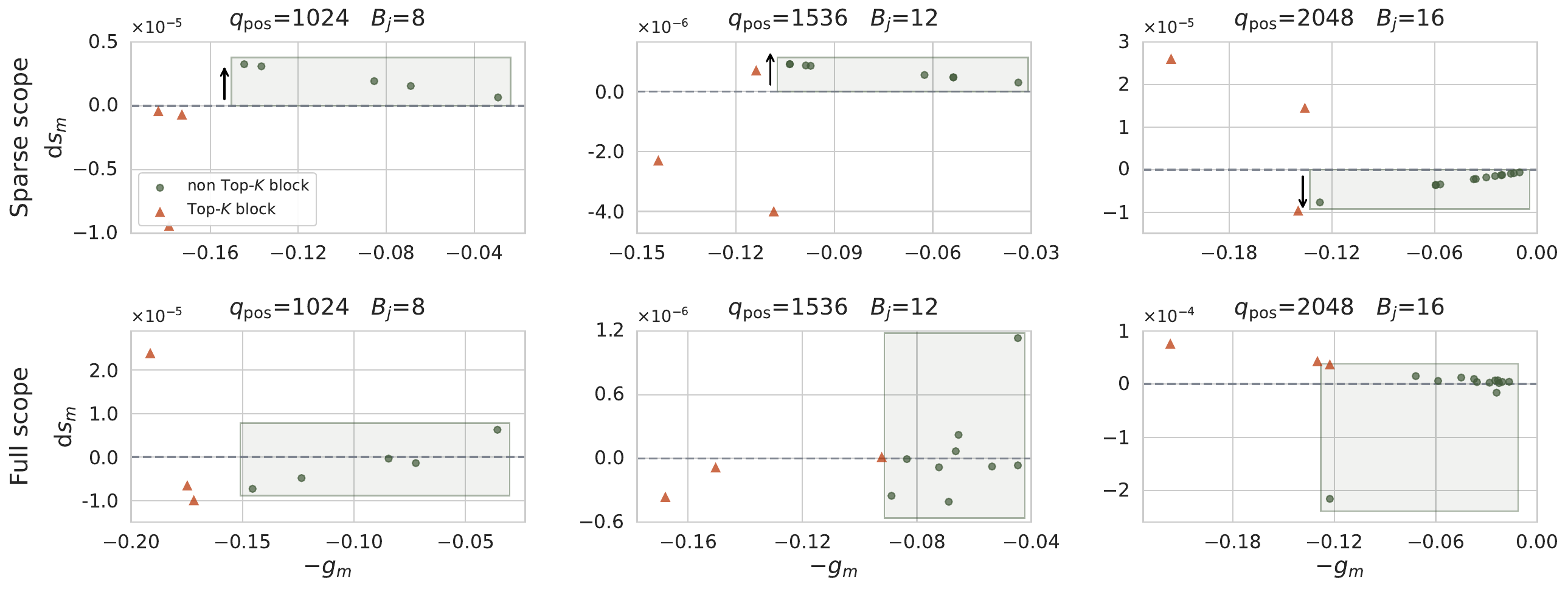}
    \caption{Gradient behavior under sparse and full training scopes with Top-$K$ of 3.
For non Top-$K$ blocks, sparse scope training yields noisy and highly correlated gradients, with gradients of these unselected blocks simultaneously being either positive or negative across blocks, ignoring their own importance scores ($x$-axis). This provides a less informative optimization signal and leads to less efficient gate optimization than full scope training.
    }
    \label{fig:ds_vs_softs}
\end{figure}

\paragraph{Training scope.}
The key difference is whether non-Top-$K$ blocks receive \textit{independent gradient signals}.
Under sparse scope, only the current Top-$K$ blocks participate in attention.
A block outside the selected set therefore has no direct gate gradient, even if its score may still change indirectly through the $\operatorname{softmax}$ normalization over selector scores.
This indirect update does not evaluate the content of the unselected block itself. For $\phi(\cdot)=\operatorname{softmax}(\cdot)$, consider an unselected block $B_m$ with $m\notin\mathcal{I}$.
Gradients under sparse and full scope are
\begin{equation}
ds_m^{sparse}
=
-g_m
\cdot
\underbrace{\sum_{\ell\in\mathcal{I}} g_\ell dg_\ell}_{\text{Top-$K$ blocks}},
\qquad
ds_m^{full}
=
-g_m
\cdot
\underbrace{
\bigg(
-dg_m
+
\sum_{\ell=1}^{C} g_\ell dg_\ell
\bigg)}_{\text{own signal + full normalization}}.
\label{eq:scope_grad}
\end{equation}
As shown in \Cref{fig:ds_vs_softs}, sparse scope updates an unselected block only through selected blocks, while full scope gives it its own  gradient $dg_m$, allowing its score to be adjusted based on its own content.

Empirically, as shown by settings (9) and (10) in \Cref{tab:ablation_main}, sparse scope starts worse than full scope but gradually catches up, reaching comparable final performance. This trend holds across model scales and budgets (Appendix~\ref{app:scope_final}).
Given its lower training cost, we therefore rely mainly on sparse scope in our subsequent experiments.
We provide the whole gradient derivations in Appendix 
\ref{app:derivation}.

\subsection{Instantiating Simple Attention Sparsification}
\label{sec:ranksparse}

Based on the above analysis, \our adopts four design choices: \emph{1)} inner $\operatorname{softmax}$ gate injection, \emph{2)} $\operatorname{softmax}$ gate activation, \emph{3)} soft gate to keep ranking information, \emph{4)} sparse scope during training.
Given the query $\mathbf{q}$, the selector computes relevance scores for the historical blocks
$\mathbf{s}
=
\mathcal{R}_{\theta}
\left(
\mathbf{q},
\{\mathbf{K}_{B_m}\}_{m=1}^{C}
\right).$
The scores are converted into normalized gates $\mathbf{g}=\operatorname{softmax}(\mathbf{s})$, and the Top-$K$ historical blocks are selected as $\mathcal{I}=\text{Top-}K(\mathbf{g},K)$. We define the attended token set as $\mathcal{S}=B_0\cup\bigcup_{m\in\mathcal{I}}B_m$, including the always-retained current block $B_0$, and broadcast the selected block gates to the token-level gate $\mathbf{g}_{\mathcal{S}}$ aligned with the attention logits. The resulting attention computation used during training is given by
\begin{equation}
\mathbf{o}_{\our}
=
\operatorname{softmax}
\left(
    \mathbf{q}\mathbf{K}^{\top}_{\mathcal{S}}
    +
    \log\mathbf{g}_{\mathcal{S}}
\right)
\mathbf{V}_{\mathcal{S}}.
\label{eq:ours_train}
\end{equation}
Thus, the selected historical blocks receive normalized log-gate biases, whereas the current block remains unbiased with a unit gate, which instantiates the differentiable continuous gating illustrated in \Cref{fig:motivation}b.
At inference time, the learned ranking is discretized into Top-$K$ block indices.
Algorithms~\ref{alg:ours_train} and~\ref{alg:ours_infer} summarize the training and inference procedures, respectively.

\begin{figure}[!t]
\centering
\newcommand{\algstrut}{\rule[-1.0ex]{0pt}{3.0ex}}
\begin{minipage}[t]{0.48\textwidth}
\begin{algorithm}[H]
\caption{\our Training Step}
\label{alg:ours_train}
\begin{algorithmic}[1]
\Require \algstrut $\mathbf{q},\mathbf{K},\mathbf{V}$, selector $\mathcal{R}_{\theta}$, budget $K$
\State \algstrut $\mathbf{s}=\mathcal{R}_{\theta}\left(\mathbf{q},\{\mathbf{K}_{B_m}\}_{m=1}^{C}\right)$
\State \algstrut $\mathbf{g}=\operatorname{softmax}(\mathbf{s})$
\State \algstrut $\mathcal{I}=\text{Top-}K(\mathbf{g},K)$, $\mathcal{S}=B_0\cup\bigcup_{m\in\mathcal{I}}B_m$
\State \algstrut $\mathbf{o}=\operatorname{softmax}(\mathbf{q}\mathbf{K}^{\top}_{\mathcal{S}}+\log\mathbf{g}_{\mathcal{S}})\mathbf{V}_{\mathcal{S}}$
\State \algstrut Update $\theta$ with $\nabla_{\theta}\mathcal{L}_{\mathrm{LM}}$
\State \algstrut \Return $\mathcal{R}_{\theta}$
\end{algorithmic}
\end{algorithm}
\end{minipage}
\hfill
\begin{minipage}[t]{0.48\textwidth}
\begin{algorithm}[H]
\caption{\our Inference Step}
\label{alg:ours_infer}
\begin{algorithmic}[1]
\Require \algstrut $\mathbf{q},\mathbf{K},\mathbf{V}$, selector $\mathcal{R}_{\theta}$, budget $K$
\State \algstrut $\mathbf{s}=\mathcal{R}_{\theta}\left(\mathbf{q},\{\mathbf{K}_{B_m}\}_{m=1}^{C}\right)$
\State \algstrut $\mathcal{I}=\text{Top-}K(\mathbf{s},K)$
\State \algstrut $\mathcal{S}=B_0\cup\bigcup_{m\in\mathcal{I}}B_m$
\State \algstrut $\mathbf{o}=\operatorname{softmax}\left(\mathbf{q}\mathbf{K}_{\mathcal{S}}^{\top}\right)\mathbf{V}_{\mathcal{S}}$
\State \algstrut Forward to obtain next token logits
\State \algstrut \Return next token
\end{algorithmic}
\end{algorithm}
\end{minipage}
\label{fig:ours_algorithm}
\end{figure}

\paragraph{Kernel Design.}
SAS training computes sparse scope attention, $\operatorname{softmax}(\mathbf{q}\mathbf{K}^\top_{\mathcal{S}}+\log\mathbf{g}_{\mathcal{S}})\mathbf{V}_{\mathcal{S}}$, where $\log\mathbf{g}_{\mathcal{S}}$ contains normalized log-gate biases for selected historical blocks and zero bias for the always-retained current block.
The selected historical set is encoded by a per-query threshold on the gate: a block is active when its log gate reaches the threshold, which realizes the Top-$K$ selection $\mathcal{I}$ without an explicit sort.
FlashAttention \cite{dao2022flashattention} kernels neither support injecting such per-block gates inside the tiled $\operatorname{softmax}$ nor skipping blocks according to a learned gate.
A naive implementation materializes the gated attention matrix, incurring prohibitive memory overhead.

We implement a FlashAttention-style Triton \cite{tillet2019triton} kernel that fuses the block gate into the tile-level $\mathbf{q}\mathbf{K}_{\mathcal{S}}^\top$ computation.
During the streaming scan over key--value tiles, the kernel adds the normalized log gate to the attention logits of each selected historical block, masks out non-selected historical blocks, leaves the always-retained current block unbiased, and performs the standard online $\operatorname{softmax}$ update.
The backward pass accumulates block-level log-gate gradients by summing the attention-logit gradients within each selected historical block.
This kernel keeps the sparse training scope while avoiding attention-matrix materialization and retaining FlashAttention-style memory efficiency. Pseudocode is provided in Appendix~\ref{sec:app_kernel}.

\section{Experiments}
\label{sec:experiments}

We primarily evaluate \our in the \emph{post-training sparsification} setting (Section~\ref{sec:exp_posttrain}), where the backbone LLM is frozen and only the selector is trained, allowing a controlled comparison against distillation-based methods. We then verify that the same end-to-end formulation extends beyond this setting to \emph{continued pretraining} (Section~\ref{sec:exp_cpt}), where the backbone and the selector are trained jointly.

\subsection{Experimental Setup}

\paragraph{Baselines, models, and evaluation.}
We first compare our method with full attention and  SeerAttention-R~\cite{gao2025seerattention}, 
which is a strong post-training sparse attention baseline that freezes the backbone LLM, uses a carefully designed AttnGate selector, and trains through dense attention distillation. 
We keep the same selector structure, but train block selection from the language modeling loss.
On reasoning tasks, we also compare with Sliding Window Attention, StreamingLLM \cite{xiao2024efficient}, and Quest \cite{tang2024quest}.
We conduct experiments on Qwen3~\cite{yang2025qwen3technicalreport} backbones (4B, 8B, 14B).
We evaluate on three types of tasks: reasoning, including MATH500~\cite{math500}, GPQA-Diamond~\cite{gpqa}, AIME24, and AIME25~\cite{aime}; long-context understanding, using LongBench~\cite{longbench}; and agentic tasks, using BFCL (Multi-Turn) \cite{patil2025bfcl} and VitaBench \cite{he2026vitabench}.
For reasoning tasks, we follow the same evaluation protocol as SeerAttention-R.
For long-context understanding, we evaluate LongBench in the non-thinking mode.
For agentic tasks, we use YaRN~\cite{peng2024yarn} to extend the context window to 65{,}536 during evaluation.

\paragraph{Training setup.}
In all settings we train only the AttnGate selector with the LLM backbone frozen, using AdamW~\cite{loshchilov2017decoupled} with learning rate 1e-3, cosine decay, and block size 64.
The selector is trained on OpenR1-MATH-220K\footnote{\url{https://huggingface.co/datasets/open-r1/OpenR1-Math-220k}} for one epoch with global batch size 32 and a maximum sequence length of 32{,}768, and the same trained selector is used across reasoning, long-context, and agentic evaluations.
Our implementation builds on VeOmni~\cite{ma2025veomni} with FSDP distributed training.

\paragraph{Inference implementation.}
We implement our method as a native attention backend in SGLang~\cite{zheng2024sglang},
built on top of its paged KV cache and FlashInfer~\cite{ye2025flashinfer} attention
kernels. Following the standard prefill/decode pipeline of LLM serving, the prefill
phase uses dense attention, while the decode phase uses block sparse attention. At
each decoding step, the trained selector scores all cached blocks, selects the
Top-$K$ blocks for each query group, and gathers only the corresponding KV blocks
for the FlashInfer sparse decode kernel. This reduces the decoding complexity from
$O(n)$ to $O(|\mathcal{S}|)$, independent of the total context length $n$. The
backend supports grouped query attention (GQA) with per-group block selection,
CUDA graph capture for the decode path, and batched concurrent decoding.
Since our method and SeerAttention-R share the same
AttnGate architecture and inference-time block selection procedure, differing only
in how the gate is trained, both run through the same backend by simply loading
their respective gate checkpoints; the dense baseline uses the same FlashInfer
paged-attention implementation on an identical model configuration.

\begin{table}[!t]
    \centering
    \caption{Evaluation results on reasoning benchmarks.
Quest$^\ast$ denotes results extracted from Figure 5 of SeerAttention-R using the WebPlotDigitizer tool, while SeerAttn-R$^\#$ denotes our reproduced results.
Standard deviations are shown in parentheses.
    }
    \label{tab:main_results}
    \resizebox{\textwidth}{!}{
    \begin{tabular}{ll lll lll lll lll}
    \toprule
    \multirow{2}{*}{\textbf{Budget}} & \multirow{2}{*}{\textbf{Method}} &
    \multicolumn{3}{c}{\textbf{MATH500}} &
    \multicolumn{3}{c}{\textbf{GPQA-Diamond}} &
    \multicolumn{3}{c}{\textbf{AIME24}} &
    \multicolumn{3}{c}{\textbf{AIME25}} \\
    \cmidrule(lr){3-5} \cmidrule(lr){6-8} \cmidrule(lr){9-11} \cmidrule(lr){12-14}
    & & \textbf{4B} & \textbf{8B} & \textbf{14B}
      & \textbf{4B} & \textbf{8B} & \textbf{14B}
      & \textbf{4B} & \textbf{8B} & \textbf{14B}
      & \textbf{4B} & \textbf{8B} & \textbf{14B} \\
    \midrule
    \textit{Full}
    & Full Attn
    & 93.93 & 94.43 & 95.22
    & 56.19 & 60.54 & 65.25
    & 71.25 & 74.48 & 78.91
    & 66.41 & 67.86 & 70.21 \\
    \midrule
    \multirow{5}{*}{1024}
    & Sliding Window
    & \valstd{20.18}{1.3} & - & -
    & \valstd{1.52\phantom{1}}{0.5}  & - & -
    & - & - & -
    & - & - & - \\
    & Quest$^*$
    & 10.38 & 31.65 & 54.45
    & 4.05  & 8.76  & 18.15
    & - & - & -
    & - & - & - \\
    & StreamingLLM
    & \valstd{72.72}{1.8} & \valstd{72.17}{1.8} & \valstd{73.47}{1.8}
    & \valstd{13.86}{1.7} & \valstd{13.04}{1.8} & \valstd{19.07}{2.2}
    & - & - & -
    & - & - & - \\
    & SeerAttn-R
    & {84.67} & {83.57} & {86.12}
    & {39.84} & {39.43} & {45.64}
    & - & - & -
    & - & - & - \\
    & \our
    & \bvalstd{90.65}{1.1}
    & \bvalstd{91.27}{1.1}
    & \bvalstd{92.93}{1.0}
    & \bvalstd{50.41}{2.8}
    & \bvalstd{53.17}{2.9}
    & \bvalstd{61.14}{2.9}
    & - & - & -
    & - & - & - \\
    \midrule
    \multirow{5}{*}{2048}
    & Sliding Window
    & \valstd{50.72}{1.8} & - & -
    & \valstd{8.93\phantom{1}}{1.6}  & - & -
    & \valstd{5.47\phantom{1}}{2.4}  & - & -
    & \valstd{2.19\phantom{1}}{1.1}  & - & - \\
    & Quest
    & 40.74 & 68.66 & 81.52
    & 12.17 & 24.83 & 41.01
    & 0     & 12.98 & 25.12
    & 0     & 12.66 & 27.22 \\
    & StreamingLLM
    & \valstd{83.65}{1.5} & \valstd{83.40}{1.5} & \valstd{84.97}{1.4}
    & \valstd{27.18}{2.4} & \valstd{27.46}{2.4} & \valstd{34.06}{2.7}
    & \valstd{29.69}{7.2} & \valstd{25.89}{7.1} & \valstd{28.85}{7.4}
    & \valstd{15.94}{6.1} & \valstd{15.94}{6.3} & \valstd{15.68}{6.2} \\
    & SeerAttn-R
    & {91.85} & {91.67} & {93.02}
    & {49.94} & {54.41} & {61.68}
    & {55.83} & {58.23} & {63.65}
    & {45.16} & {43.30} & {48.70} \\
    & \our
    & \bvalstd{93.47}{1.0}
    & \bvalstd{93.17}{1.0}
    & \bvalstd{93.54}{1.0}
    & \bvalstd{54.86}{2.8}
    & \bvalstd{58.74}{2.8}
    & \bvalstd{65.09}{2.8}
    & \bvalstd{68.85}{6.6}
    & \bvalstd{70.73}{6.7}
    & \bvalstd{77.76}{6.6}
    & \bvalstd{56.38}{7.3}
    & \bvalstd{58.41}{7.2}
    & \bvalstd{64.19}{7.1} \\
    \midrule
    \multirow{6}{*}{4096}
    & Sliding Window
    & \valstd{75.33}{1.6} & - & -
    & \valstd{23.74}{2.6} & - & -
    & \valstd{19.74}{5.7} & - & -
    & \valstd{16.35}{5.7} & - & - \\
    & Quest
    & 71.59 & 86.68 & 90.76
    & 23.21 & 43.18 & 58.02
    & 12.41 & 43.99 & 46.67
    & 12.41 & 32.14 & 42.13 \\
    & StreamingLLM
    & \valstd{91.03}{1.1} & \valstd{91.12}{1.1} & \valstd{92.07}{1.1}
    & \valstd{40.40}{2.8} & \valstd{42.27}{2.8} & \valstd{47.35}{2.9}
    & \valstd{45.94}{7.9} & \valstd{46.35}{8.0} & \valstd{49.32}{8.1}
    & \valstd{31.56}{7.3} & \valstd{31.15}{7.4} & \valstd{32.60}{7.5} \\
    & SeerAttn-R
    & \textbf{94.10} & 94.00 & {95.12}
    & \textbf{55.40} & \textbf{60.48} & {63.83}
    & {69.32} & {71.35} & {75.73}
    & {58.59} & {57.81} & {64.79} \\
    & SeerAttn-R$^{\#}$
    & \valstd{93.00}{1.1}
    & \bvalstd{94.25}{0.9} & \valstd{94.83}{0.8}
    & \valstd{54.86}{2.8}
    & \valstd{59.03}{2.9} & \valstd{63.67}{2.8}
    & \valstd{69.16}{6.9}
    & \valstd{69.43}{6.9} & \valstd{75.10}{6.8}
    & \valstd{56.59}{7.6}
    & \valstd{57.66}{7.4} & \valstd{64.69}{7.0} \\
    & \our
    & \valstd{93.67}{1.0}
    & \valstd{94.23}{1.0}
    & \bvalstd{95.23}{0.8}
    & \valstd{55.11}{2.9}
    & \valstd{60.42}{2.8}
    & \bvalstd{65.03}{2.8}
    & \bvalstd{71.72}{6.8}
    & \bvalstd{73.49}{6.7}
    & \bvalstd{78.28}{6.4}
    & \bvalstd{59.97}{7.3}
    & \bvalstd{63.41}{7.2}
    & \bvalstd{67.29}{6.9} \\
    \bottomrule
    \end{tabular}
    }
\end{table}

\begin{table}[!t]
    \centering
    \caption{Evaluation results on LongBench, grouped by input length (0-4K, 4-8K, and 8K+ tokens).}
    \label{tab:longc}
    \resizebox{0.95\textwidth}{!}{
    \begin{tabular}{ll cccc cccc cccc}
    \toprule
    \multirow{2}{*}{\textbf{Budget}} & \multirow{2}{*}{\textbf{Method}}
    & \multicolumn{4}{c}{\textbf{Qwen3-4B}}
    & \multicolumn{4}{c}{\textbf{Qwen3-8B}}
    & \multicolumn{4}{c}{\textbf{Qwen3-14B}}  \\
    \cmidrule(lr){3-6} \cmidrule(lr){7-10} \cmidrule(lr){11-14}
    & & 0-4K & 4-8K & 8K+ & Avg.
      & 0-4K & 4-8K & 8K+ & Avg.
      & 0-4K & 4-8K & 8K+ & Avg. \\
    \midrule
    \textit{Full}
    & Full Attn
    & 53.9 & 52.4 & 50.5 & 52.2
    & 57.4 & 54.1 & 51.7 & 54.4
    & 59.2 & 55.6 & 55.0 & 56.6 \\
    \midrule

    2048
    & SeerAttn-R
    & 53.5 & 51.6 & 48.4 & 51.2
    & \textbf{57.0} & 53.1 & 47.6 & 52.6
    & 58.9 & 54.3 & 51.5 & 54.9 \\

    & \our
    & \textbf{53.6} & \textbf{52.1} & \textbf{48.8} & \textbf{51.5}
    & \textbf{57.0} & \textbf{53.9} & \textbf{49.3} & \textbf{53.4}
    & \textbf{59.2} & \textbf{54.4} & \textbf{53.9} & \textbf{55.8} \\

    \midrule

    4096
    & SeerAttn-R
    & \textbf{54.0} & 52.1 & 49.7 & 51.9
    & \textbf{57.4} & 53.9 & 50.0 & 53.8
    & \textbf{59.0} & \textbf{54.8} & 53.5 & 55.7 \\

    & \our
    & 53.8 & \textbf{52.4} & \textbf{49.8} & \textbf{52.0}
    & \textbf{57.4} & \textbf{54.1} & \textbf{50.4} & \textbf{53.9}
    & 58.9 & \textbf{54.8} & \textbf{54.8} & \textbf{56.2}\\

    \bottomrule
    \end{tabular}
    }
\end{table}

\begin{table}[!t]
    \centering
    \caption{Evaluation results on BFCL (Multi-Turn).}
    \label{tab:bfcl}

    \vspace{2pt}
    \resizebox{0.6\linewidth}{!}{
    \begin{tabular}{llccc}
    \toprule
    \textbf{Budget} & \textbf{Method}
    & \textbf{Qwen3-4B}
    & \textbf{Qwen3-8B}
    & \textbf{Qwen3-14B}
    \\
    \midrule

    \textit{Full}
    & Full Attn
    & 35.75
    & 42.75
    & 44.50
    \\

    \midrule

    2048
    & SeerAttn-R
    & 29.00
    & 34.38
    & 38.88
    \\
    & \our
    & \textbf{32.50}
    & \textbf{36.75}
    & \textbf{39.50}
    \\

    \midrule

    4096
    & SeerAttn-R
    & 33.38
    & 39.25
    & 43.88
    \\
    & \our
    & \textbf{34.38}
    & \textbf{41.25}
    & \textbf{44.00}
    \\

    \bottomrule
    \end{tabular}
    }
\end{table}

\begin{table}[!t]
    \centering
    \caption{Evaluation results on VitaBench with Qwen3-14B.}
    \label{tab:vitabench}

    \vspace{2pt}
    \resizebox{0.98\linewidth}{!}{
    \begin{tabular}{llrrrrrrrrr}
    \toprule
    \multirow{2}{*}{\textbf{Budget}} &
    \multirow{2}{*}{\textbf{Method}}
    & \multicolumn{3}{c}{\textbf{Delivery}}
    & \multicolumn{3}{c}{\textbf{Instore}}
    & \multicolumn{3}{c}{\textbf{OTA}}
    \\
    \cmidrule(lr){3-5}
    \cmidrule(lr){6-8}
    \cmidrule(lr){9-11}
    & & Avg@4 & Pass@4 & Pass\textasciicircum{}{4}
    & Avg@4 & Pass@4 & Pass\textasciicircum{}{4}
    & Avg@4 & Pass@4 & Pass\textasciicircum{}{4}\\
    \midrule

    \textit{Full}
    & Full Attn
    & \valstd{29.0}{3.3}
    & \valstd{62.0}{4.8}
    & \valstd{8.0}{2.7}
    & \valstd{27.2}{2.9}
    & \valstd{64.0}{4.8}
    & \valstd{5.0}{2.2}
    & \valstd{12.5}{2.1}
    & \valstd{37.0}{4.8}
    & \valstd{1.0}{1.0}
    \\

    \midrule

    2048
    & SeerAttn-R
    & \bvalstd{32.2}{3.0}
    & \bvalstd{67.0}{4.7}
    & \valstd{4.0}{2.0}
    & \valstd{19.0}{2.5}
    & \valstd{48.4}{5.1}
    & \valstd{2.1}{1.5}
    & \valstd{4.2}{1.4}
    & \valstd{14.9}{4.3}
    & \valstd{0.0}{0.0}
    \\

    & \our
    & \valstd{30.0}{3.3}
    & \valstd{61.0}{4.9}
    & \bvalstd{9.0}{2.9}
    & \bvalstd{19.2}{2.5}
    & \bvalstd{51.0}{5.0}
    & \bvalstd{3.0}{1.7}
    & \bvalstd{10.2}{1.8}
    & \bvalstd{28.0}{4.5}
    & \valstd{0.0}{0.0}
    \\

    \midrule

    4096
    & SeerAttn-R
    & \valstd{32.5}{3.1}
    & \valstd{63.0}{4.8}
    & \bvalstd{4.0}{2.0}
    & \valstd{26.7}{2.8}
    & \valstd{56.0}{4.9}
    & \valstd{1.0}{1.0}
    & \valstd{11.7}{2.2}
    & \bvalstd{29.0}{4.5}
    & \valstd{1.0}{1.0}
    \\

    & \our
    & \bvalstd{34.2}{3.0}
    & \bvalstd{68.0}{4.6}
    & \bvalstd{4.0}{2.0}
    & \bvalstd{28.7}{3.0}
    & \bvalstd{61.0}{4.9}
    & \bvalstd{5.0}{2.2}
    & \bvalstd{12.3}{2.3}
    & \valstd{27.0}{4.4}
    & \bvalstd{2.0}{1.4}
    \\

    \bottomrule
    \end{tabular}
    }
\end{table}

\subsection{Post-Training Results}
\label{sec:exp_posttrain}

\paragraph{Reasoning task performance.}
As shown in \Cref{tab:main_results}, \our{} outperforms all sparse baselines across nearly all budgets, backbones, and tasks.
The training-free static-sparsity baselines (Sliding Window, StreamingLLM) and the query-aware Quest all degrade sharply at tight budgets: the fixed local (plus optional sink) patterns of Sliding Window and StreamingLLM discard reasoning-critical context, and Quest collapses entirely (e.g., 0 on AIME24/25 at budget 2048), whereas \our{} stays strong.
Against SeerAttention-R, which uses the identical AttnGate but is trained by attention distillation, \our{} still wins clearly on harder tasks (e.g., +13.0 on AIME24 at budget 2048, Qwen3-4B); since the two differ only in the training signal, this directly shows the language modeling loss is a more effective supervision for sparse selection.
At budget 4096, \our{} matches or exceeds full attention in several cases (e.g., 71.72 vs.\ 71.25 on AIME24, Qwen3-4B), recovering full attention reasoning while attending to only a small fraction of the KV blocks.

\paragraph{Long context understanding performance.}
\Cref{tab:longc} reports LongBench results, where the selector is trained only on math data and evaluated directly on long-context inputs.
Despite this distribution shift, \our{} outperforms SeerAttention-R at nearly every budget and backbone, with the gap widening on longer inputs where selection quality matters most: at budget 2048 on Qwen3-14B, \our{} improves the 8K+ bucket by +2.4 (53.9 vs.\ 51.5).
At budget 4096, \our{} essentially recovers full attention (e.g., 56.2 vs.\ 56.6 average on Qwen3-14B), showing that a selector trained purely on math data still transfers effectively to long-context understanding.

\paragraph{Agentic task performance.}
On BFCL multi-turn (\Cref{tab:bfcl}), \our{} beats SeerAttention-R across all backbones and budgets, e.g., +3.5 on Qwen3-4B at budget 2048, and nearly closes the gap to full attention at budget 4096 (e.g., 44.00 vs.\ 44.50 on Qwen3-14B).
The advantage carries over to VitaBench (\Cref{tab:vitabench}): at budget 4096, \our{} leads SeerAttention-R on most metrics across the Delivery, Instore, and OTA scenarios and approaches full attention, showing that end-to-end selection remains reliable in realistic long-horizon tool-use settings.

\subsection{Extension to Continued Pretraining}
\label{sec:exp_cpt}

\paragraph{Baselines, model, and evaluation.}
We follow HiLS-Attention~\cite{hu2026hierarchical} to conduct a continued pretraining experiment.
We start from the OLMo3-7B~\cite{olmo2026olmo3} stage-1 base checkpoint and apply \our with block size 64 and Top-$K$ 32.
Instead of the AttnGate used in post-training, here we design a lightweight selector with a dedicated retrieval query/key projection, initialized from the attention projections. All query heads within a group share the same selected blocks.
We compare with three baselines built on the same checkpoint: \emph{OLMo3-Base}, the original dense model; \emph{OLMo3-512SWA}, a sliding-window baseline with continued pretraining; and \emph{HiLS-Attn-RoPE}, the RoPE variant of HiLS-Attention (HiLS-Attn RoPE-Q-Cal in their paper).
We evaluate on general downstream tasks, spanning general knowledge (MMLU \cite{hendryckstest2021}, GPQA \cite{rein2024gpqa}, HellaSwag \cite{zellers2019hellaswag}, ARC-c \cite{clark2018think}, BoolQ \cite{clark2019boolq}, Race \cite{lai-etal-2017-race}), mathematics (CMath \cite{wei2023cmathlanguagemodelpass}, GSM8K \cite{cobbe2021gsm8k}), and code (CRUX \cite{gu2024cruxevalbenchmarkcodereasoning}, HumanEval+ \cite{evalplus}, MBPP+ \cite{austin2021programsynthesislargelanguage}), and on long-context understanding with LongBench.

\paragraph{Training setup.}
Here we train the whole model, both the OLMo3 backbone and the selector, jointly.
Following HiLS-Attention, we train on the OLMo3 pretraining corpus with a maximum sequence length of 8{,}192 for 13{,}000 steps with a global batch size of 512, corresponding to about 50B tokens, using AdamW with the learning rate warmed up to 2e-4 and cosine decayed to 2e-5.

\begin{table}[!t]
    \centering
    \caption{
    Evaluation results on general downstream tasks for OLMo3-7B continued pretraining.
    }
    \label{tab:cpt_general}
    \resizebox{0.75\textwidth}{!}{
    \begin{tabular}{l c|ccc}
    \toprule
    \textbf{Task} & {Olmo3-Base}
    & {Olmo3-512SWA}
    & {HiLS-Attn RoPE}
    & {\our-RoPE}
    \\
    \midrule
    \textit{General Knowledge} &&&&\\
    MMLU (5-shot) & 59.90 & \textbf{59.12} & 56.69 & 58.58 \\
    GPQA (5-shot) & 29.29 & \textbf{31.31} & 24.75 & 26.77 \\
    Hellaswag (10-shot) & 44.17 & 42.96 & 33.17 & \textbf{50.63} \\
    ARC-c (25-shot) & 53.56 & \textbf{55.59} & 54.92 & 52.54 \\
    BoolQ (5-shot)  & 61.01 & \textbf{64.22} & 63.43 & 62.87 \\
    Race (3-shot) & 73.89 & 72.97 & 69.50 & \textbf{74.05} \\
    \midrule
    \textit{Mathematics} &&&&\\
    CMath & 41.53 & 39.98 & \textbf{42.44} & 42.17 \\
    GSM8K & 37.00 & 33.43 & \textbf{35.71} & 34.42 \\
    \midrule
    \textit{Code} &&&&\\
    CRUX & 24.62 & 24.50 & \textbf{25.62} & 19.25 \\
    HumanEval+ & 20.10 & 19.50 & 18.90 & \textbf{20.10} \\
    MBPP+ & 37.60 & 32.30 & 33.30 & \textbf{34.60} \\
    \midrule
    Average & 43.88 & 43.24 & 41.68 & \textbf{43.28} \\
    \bottomrule
    \end{tabular}
    }
\end{table}

\begin{table}[!t]
    \centering
    \caption{
        Evaluation results on LongBench for OLMo3-7B continued pretraining.
    }
    \label{tab:cpt_longbench}
    \resizebox{\textwidth}{!}{
    \begin{tabular}{l cccccc cccccc c}
    \toprule
    \multirow{2}{*}{\textbf{Model}}
    & \multicolumn{6}{c}{\textbf{$<$8K}}
    & \multicolumn{6}{c}{\textbf{$>$8K}}
    & \multirow{2}{*}{\textbf{Average}} \\
    \cmidrule(lr){2-7} \cmidrule(lr){8-13}
    & SDoc & MDoc & Summ. & Few-shot & Synth. & Code
    & SDoc & MDoc & Summ. & Few-shot & Synth. & Code & \\
    \midrule
    Olmo3-Base
    & 36.9 & 33.9 & 23.4 & 63.8 & 5.8 & 61.8
    & 11.4 & 14.4 & 12.1 & 36.0 & 3.3 & 41.0 & 29.0 \\
    \midrule
    Olmo3-512SWA
    & \textbf{37.4} & 27.8 & 22.1 & \textbf{64.1} & 4.1 & 59.4
    & 10.0 & 14.2 & 12.9 & 33.6 & 3.2 & 32.9 & 28.0 \\
    HiLS-Attn-RoPE
    & 37.2 & 33.1 & 22.2 & 61.0 & \textbf{5.4} & \textbf{60.6}
    & \textbf{17.5} & \textbf{17.3} & 14.0 & \textbf{42.0} & \textbf{4.3} & 47.1 & \textbf{30.0} \\
    \our-RoPE
    & 34.4 & \textbf{34.7} & \textbf{22.8} & \textbf{64.1} & 4.1 & 58.8
    & 16.1 & 16.3 & \textbf{14.8} & 41.4 & 3.4 & \textbf{48.3} & \textbf{30.0} \\
    \bottomrule
    \end{tabular}
    }
\end{table}

\paragraph{Downstream task and long-context performance.}
\Cref{tab:cpt_general} shows that, among the continued-pretrained models, \our achieves the best average score (43.28), performing comparably to the sliding-window baseline OLMo3-512SWA (43.24), clearly ahead of HiLS-Attn-RoPE (41.68), and nearly matching the dense OLMo3-Base (43.88).
\our leads on knowledge tasks such as HellaSwag (50.63) and Race (74.05), and remains competitive on mathematics and code.
\Cref{tab:cpt_longbench} further shows that \our attains a tied-best LongBench average (30.0), clearly surpassing the dense base (29.0) and the sliding-window continued-pretraining baseline (28.0).
The gains concentrate on long inputs ($>$8K), where accurate block selection matters most, providing initial evidence that end-to-end sparse selection also transfers to the continued pretraining stage.

\section{Analysis}
\label{sec:analysis}

This section analyzes \emph{why} end-to-end training yields a better sparse selector.
We first inspect the learned block selection, and find that \our covers less attention mass per layer yet its cross-layer union better matches the full attention oracle.
We then show that this more effective selection lets \our reach answers with shorter reasoning traces and fewer truncations.
Finally, we quantify the end-to-end decode efficiency and report substantial speedups over full attention.

\subsection{Analyzing the Learned Block Selection}
\label{sec:ana_selection}

\begin{figure}[!t]
    \centering

    \begin{subfigure}{0.99\linewidth}
        \centering
        \includegraphics[width=\linewidth]{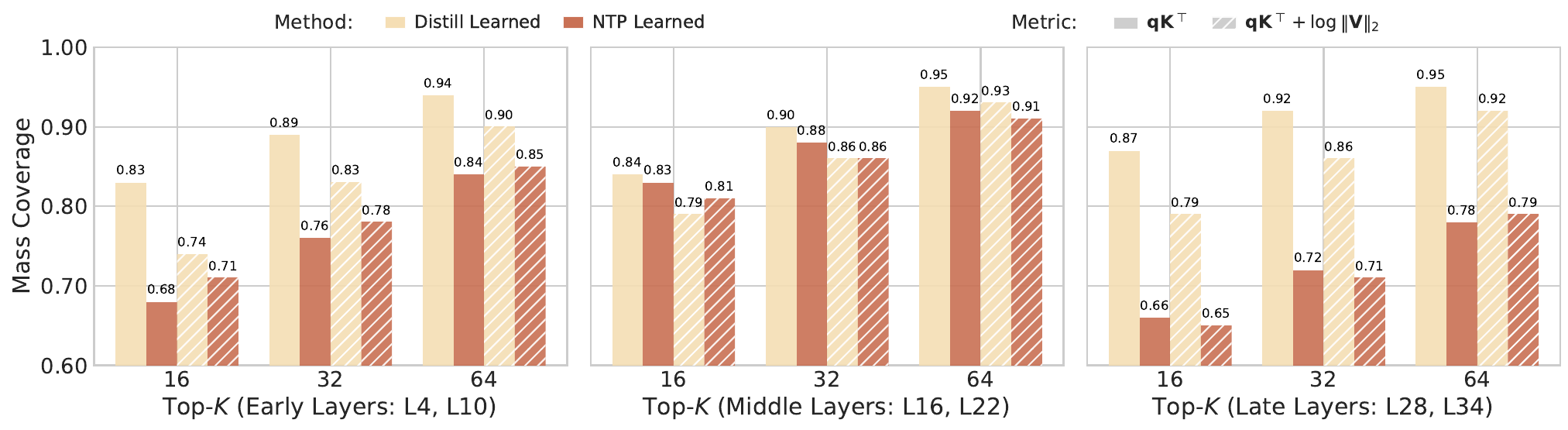}
        \caption{}
        \label{fig:mass_coverage}
    \end{subfigure}

    \begin{subfigure}{0.99\linewidth}
        \centering
        \includegraphics[width=\linewidth]{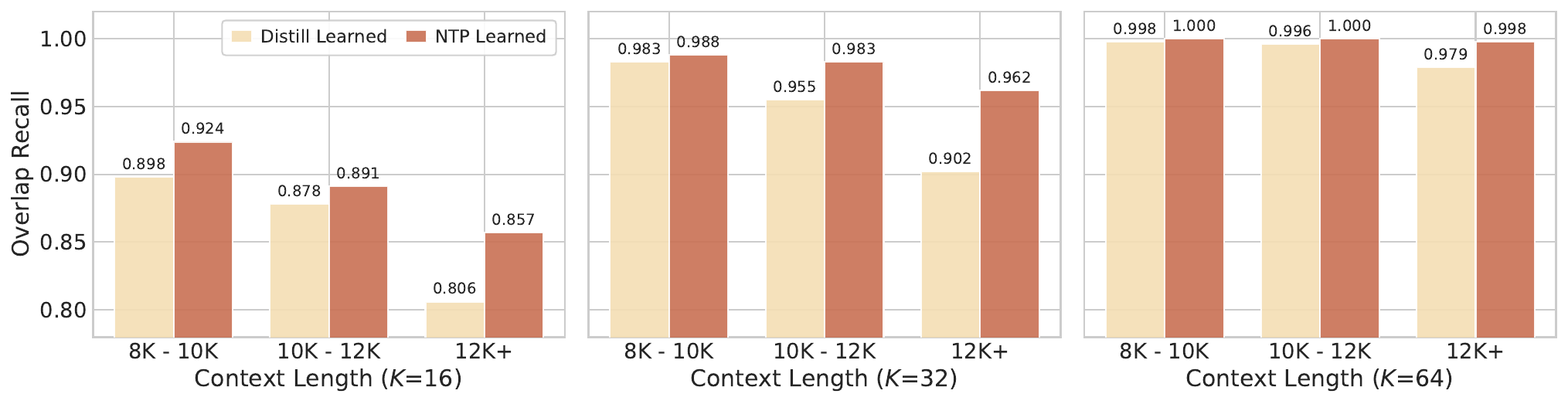}
        \caption{}
        \label{fig:overlap_recall}
    \end{subfigure}

    \caption{Analysis of the learned block selection, comparing the block selection learned by distillation (SeerAttention-R) and by end-to-end NTP (next-token prediction) training (\our).
    (a) Per-layer attention mass coverage of the selected Top-$K$ blocks, under both raw weight $\mathbf{q}\mathbf{K}^\top$, and value-aware weight $\mathbf{q}\mathbf{K}^\top+\log\|\mathbf{V}\|_2$ from STEM \cite{niu2026stemrethinkingcausalinformation}. \our covers \emph{less} mass than distillation.
    (b) Overlap recall of the cross-layer union of selected blocks against the full attention oracle. \our attains \emph{higher} recall than distillation, consistent with more complementary cross-layer selection.
    }
    \label{fig:analysis}
\end{figure}

\paragraph{Per-layer attention mass coverage.}
A natural way to assess a sparse selector is how much attention mass its selected token set $\mathcal{S}$ covers.
Reusing the attention logit $z_i=\mathbf{q}\mathbf{k}_i^\top$, we define the covered mass ratio with a per token weight $w_i$ as
\begin{equation}
    \operatorname{Cov}(\mathcal{S})
    =
    \frac{\sum_{i\in\mathcal{S}} w_i}{\sum_{j} w_j}.
    \label{eq:mass_coverage}
\end{equation}
We consider two instantiations of $w_i$. The first is $w_i=\exp(z_i)$, which depends only on the attention weight. The second is $w_i=\exp(z_i+\log\|\mathbf{v}_i\|_2)$, which additionally incorporates the magnitude of the value vector $\mathbf{v}_i$, following~\cite{niu2026stemrethinkingcausalinformation}.
As our objective is the final language modeling loss, which depends on all attention components including $\mathbf{V}$, we expect \our to favor blocks that contribute larger attention outputs.
However, \Cref{fig:mass_coverage} shows that \our covers \emph{less} mass than distillation under both metrics in almost all layers.
One possible explanation is that distillation explicitly optimizes for full attention coverage as its training target, naturally encouraging higher attention mass coverage, whereas \our is never trained to match the full attention distribution.

\paragraph{Overlap with the full attention oracle.}
The per-layer coverage above, however, ignores that selection is distributed across layers: a block missed at one layer can still be attended at another.
We therefore take, for each method, the union of its selected blocks across all layers and measure how much of the full attention model's attended blocks it recovers.
On long training sequences ($>$8K) with Qwen3-4B, we obtain the oracle set from full attention; since it uses GQA, we aggregate the per-head Top-$K$ within each group into one group-level set by majority vote.
Taking the union of these sets over all $L$ layers gives the oracle set $\mathcal{I}_{\text{all}}^{\star}$, and applying the same procedure to a method's own selection gives $\mathcal{I}_{\text{all}}$.
We then report the overlap recall
\begin{equation}
    \operatorname{Recall}(\mathcal{I}_{\text{all}}, \mathcal{I}_{\text{all}}^{\star})
    =
    \frac{|\mathcal{I}_{\text{all}}\cap\mathcal{I}_{\text{all}}^{\star}|}{|\mathcal{I}_{\text{all}}^{\star}|},
    \qquad
    \mathcal{I}_{\text{all}}=\bigcup_{j=1}^{L}\mathcal{I}_{j},
    \label{eq:overlap_recall}
\end{equation}
where $\mathcal{I}_{j}$ is the per-layer block set at layer $j$ and $\mathcal{I}_{\text{all}}^{\star}$ is defined analogously.
As shown in \Cref{fig:overlap_recall}, \our attains consistently higher overlap recall across context lengths and budgets.
This matches our expectation: distillation fits the full attention layer by layer as a local objective, whereas \our is trained end-to-end, so its block selection is optimized jointly across layers.
As a result, even though \our retains less attention mass within each individual layer, its per-layer choices are more complementary, and their union attains higher overlap recall against the oracle.

\subsection{Analyzing the Generation Behavior}
\label{sec:ana_token}

\begin{figure}[!t]
    \centering
    \includegraphics[width=0.99\linewidth]{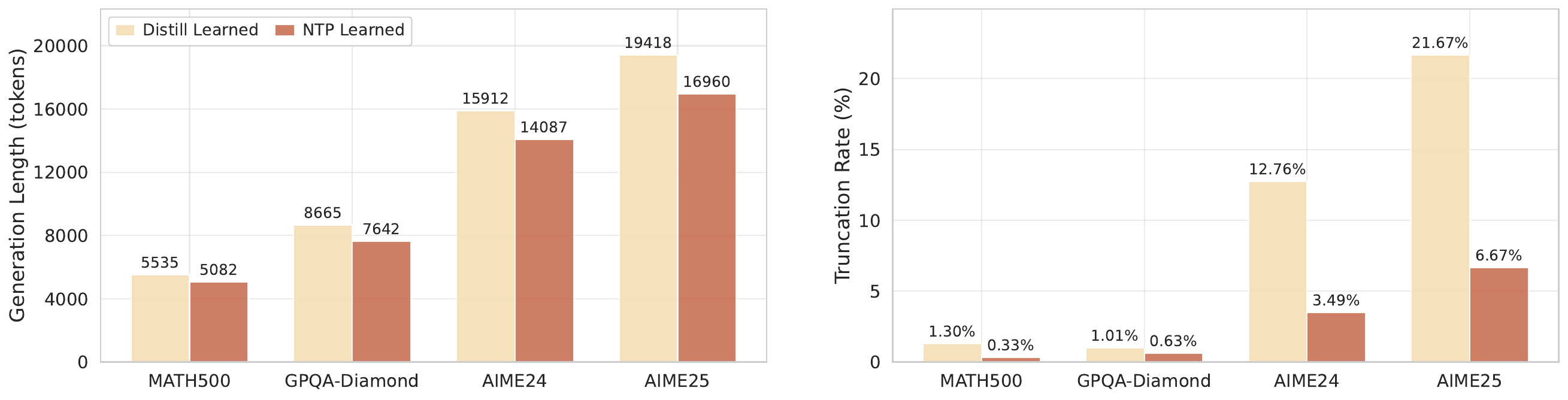}
    \caption{Generation behavior of distillation (SeerAttention-R) and end-to-end NTP (\our) on Qwen3-4B at a 4096 token budget across four reasoning benchmarks. Left: average generation length; \our generates shorter traces.
    Right: truncation rate under a maximum generation length of 32{,}768; \our truncates less often. These trends are consistent with more effective context selection, suggesting \our reaches answers with fewer tokens, with the gap largest on the harder AIME tasks.}
    \label{fig:token_eff}
\end{figure}

Beyond final accuracy, the quality of sparse selection also affects token efficiency, measured here by the generation length needed to reach an answer.
When important context is dropped, the model tends to generate longer, less focused reasoning traces and is more likely to hit the maximum generation length before finishing.
We therefore compare SeerAttention-R and \our on Qwen3-4B at a 4096 token budget, reporting the average generation length and the truncation rate under a maximum length of 32{,}768 (\Cref{fig:token_eff}).
Across all four reasoning benchmarks, \our produces shorter generations and a lower truncation rate, with the gap widening on the harder AIME tasks where long reasoning is required.
These results are consistent with more effective context selection: they suggest that \our can complete its reasoning with fewer tokens and less frequent truncation, complementing the accuracy gains in Section~\ref{sec:exp_posttrain}.

\subsection{End-to-End Decode Efficiency}
\label{sec:e2e_efficiency}

\begin{figure}[!t]
    \centering

    \begin{subfigure}{0.33\linewidth}
        \centering
        \includegraphics[width=\linewidth]{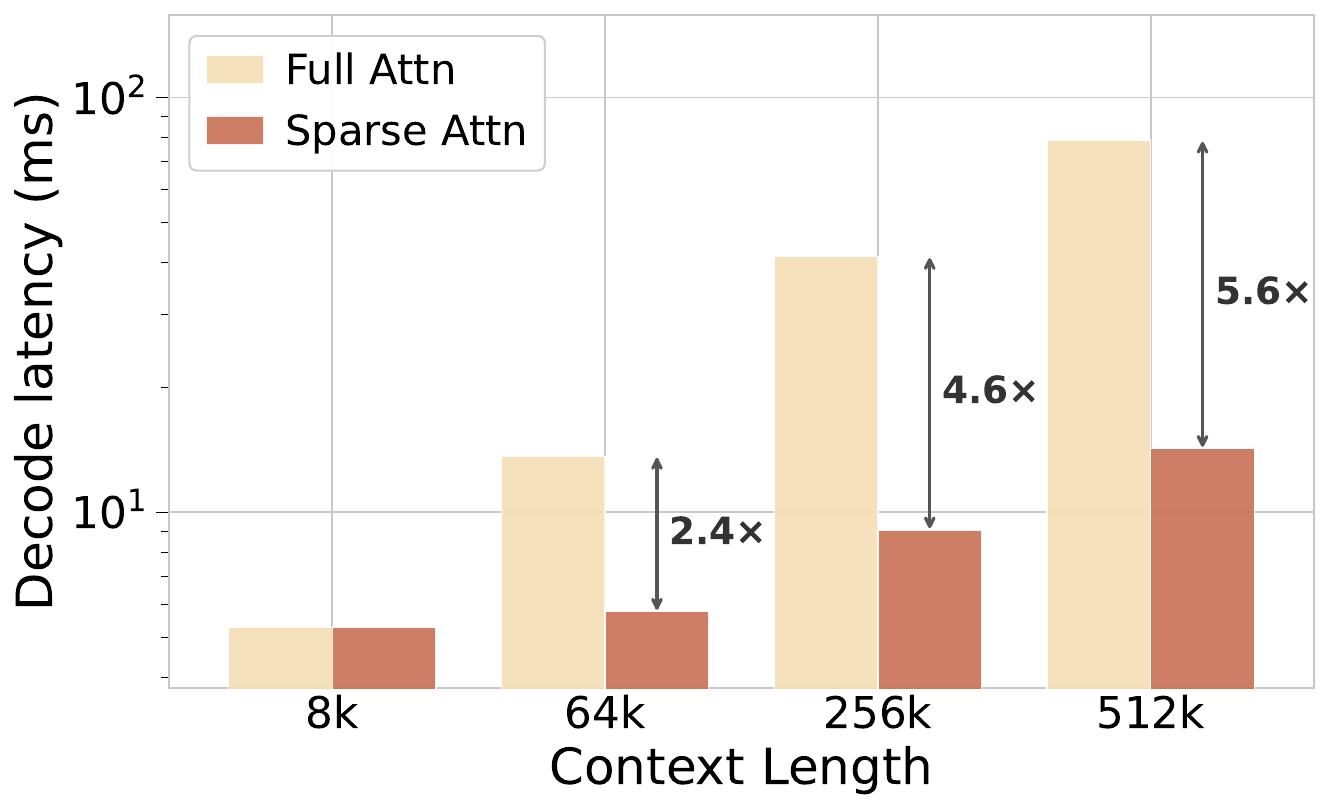}
        \caption{}
        \label{fig:e2e_latency}
    \end{subfigure}\hfill
    \begin{subfigure}{0.33\linewidth}
        \centering
        \includegraphics[width=\linewidth]{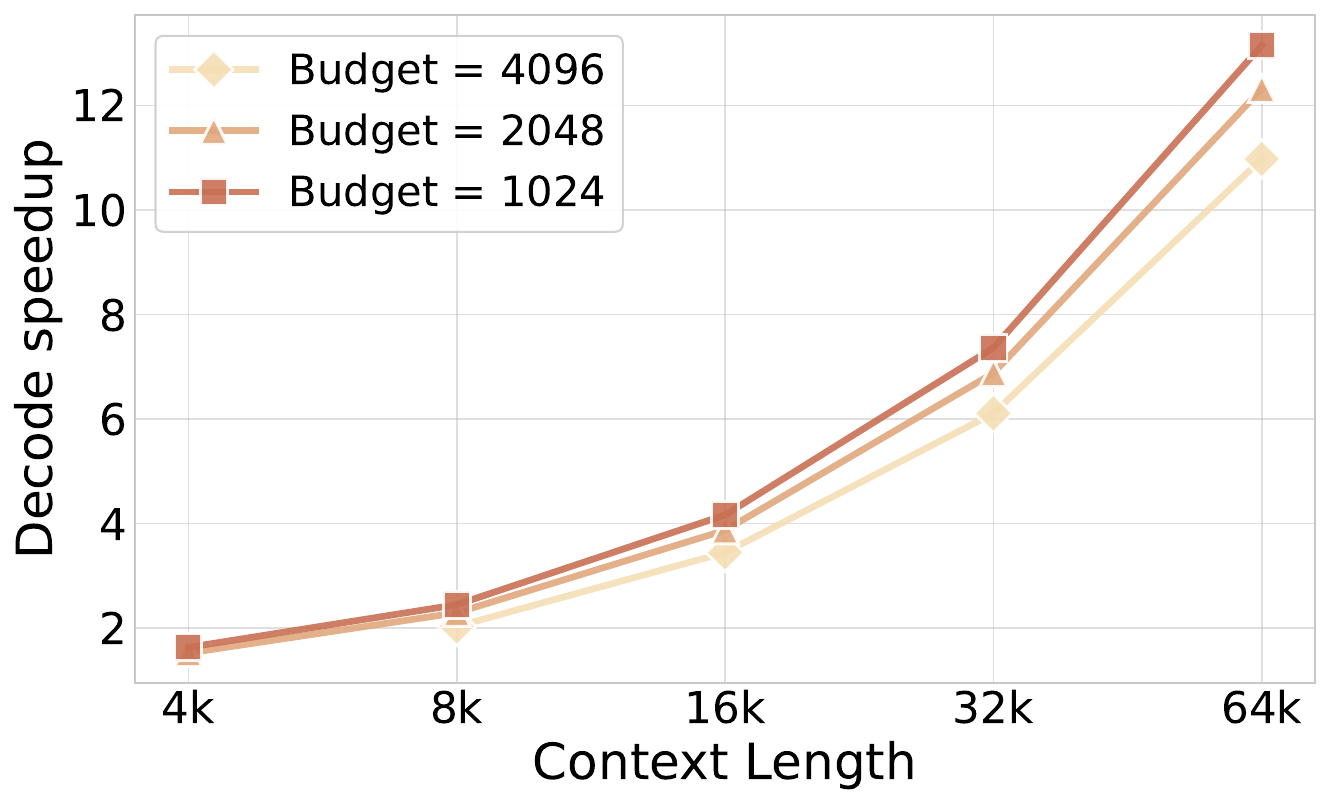}
        \caption{}
        \label{fig:e2e_budget}
    \end{subfigure}\hfill
    \begin{subfigure}{0.33\linewidth}
        \centering
        \includegraphics[width=\linewidth]{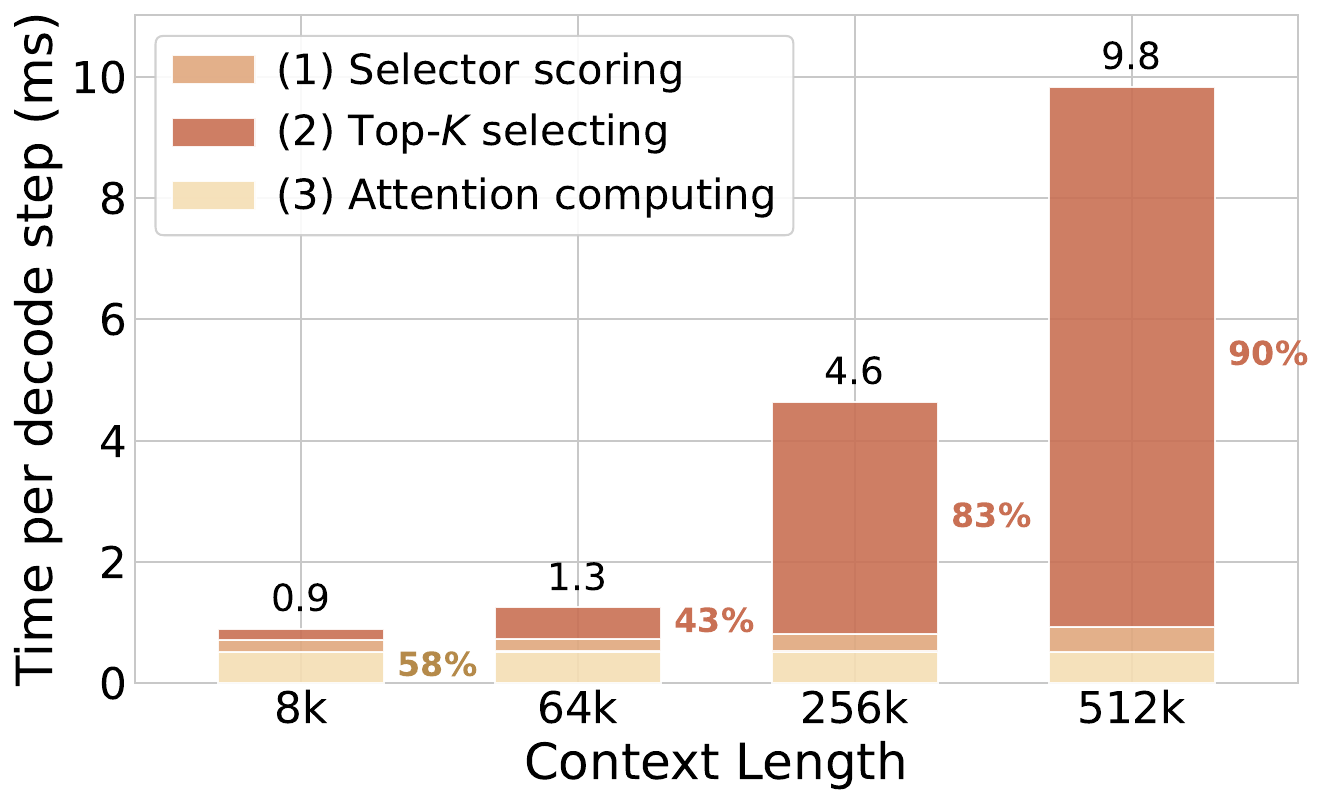}
        \caption{}
        \label{fig:e2e_breakdown}
    \end{subfigure}

    \caption{
    End-to-end decode efficiency of \our versus dense attention on Qwen3-4B in SGLang (single GPU, CUDA graphs, steady-state decode).
    (a) At batch 1, dense latency grows linearly with context length, while \our remains nearly constant, achieving up to 5.6$\times$ lower latency at 512K.
    (b) At batch 8, speedup reaches $\sim$13$\times$ and is largely insensitive to the token budget.
    (c) Sparse-decode latency breakdown: attention cost remains constant, while Top-$K$ selection dominates at long contexts.
    }
    \label{fig:e2e_efficiency}
\end{figure}

\paragraph{Decode speedup over full attention.}
We measure wall-clock decode efficiency by serving Qwen3-4B in SGLang with
our block sparse attention against full attention under an identical
configuration: a single GPU (tensor parallel size 1), CUDA graphs enabled,
chunked prefill, and a steady-state decode measurement (the input is prefilled,
then 512 generated tokens are timed).
Since \our sparsifies \emph{decode-time} attention only, prefill is unchanged,
and the reported numbers isolate the per-token decode cost.
Full attention must read the entire KV cache at every step, so its decode
latency grows linearly with context length, whereas \our reads only a fixed
token budget of blocks and stays essentially flat.
The gap therefore widens with context: at batch 1
(\Cref{fig:e2e_latency}) the two are near parity at 8K but \our is
2.4$\times$, 4.6$\times$, and 5.6$\times$ faster at 64K, 256K, and 512K.
Increasing the batch amortizes the fixed per-step overhead and further favors
sparsity: at batch 8 (\Cref{fig:e2e_budget}) the speedup reaches
$\sim$13$\times$ at 64K.
A tighter token budget reads fewer blocks and is thus faster, but the difference
across budgets is modest, indicating that the benefit
comes primarily from avoiding the full KV read rather than from the exact budget.

\paragraph{Decode step cost breakdown.}
To understand the scaling behavior, we decompose one sparse attention decode
step into its three components: selector scoring (the gate query projection and
its dot-product against the block summaries), Top-$K$ block selection, and the
attention compute over the selected blocks (\Cref{fig:e2e_breakdown}).
The attention compute is context invariant, since the number of blocks read is
capped by the token budget regardless of sequence length.
Selector scoring, by contrast, scans all block summaries and thus grows with
context in absolute terms, rising from well below the attention compute at 8K
to roughly matching it at 512K; extrapolating to longer context, it is
expected to surpass the attention compute.
The Top-$K$ selection grows even faster, as it must rank all candidate blocks,
and comes to dominate the step, shifting from 21$\%$ at 8K
(where attention compute dominates at 58$\%$) to 90$\%$ at 512K.
This identifies the selection stage, Top-$K$ ranking together with the growing
selector scoring, rather than the attention compute, as the primary bottleneck
and the main target for further kernel optimization at very long context.

\section{Conclusion}
\label{sec:conclusion}

In this paper, we presented \our, a simple yet effective paradigm for end-to-end post-training attention sparsification. By shifting from layer-wise attention distillation to direct optimization with language modeling loss, \our overcomes the surrogate supervision and captures cross-layer dependencies more effectively. 
Our empirical analysis identifies the design choices that enable stable, informative gradient flow for context selection, including log-space gate injection, normalized gate activation, and preservation of continuous rankings. Facilitated by our fused Triton kernels, \our is computationally practical for long-context training and superior in performance. 
Extensive evaluations across reasoning, long-context understanding, and agentic tasks demonstrate that \our outperforms existing baselines, particularly in budget-constrained scenarios.

\bibliographystyle{plainnat}
\bibliography{references}

\clearpage
\newpage
\appendix

\section{Training Dynamics}
\label{app:dynamics}
\begin{figure}[h]
    \centering

    \begin{subfigure}{0.99\linewidth}
        \centering
        \includegraphics[width=\linewidth]{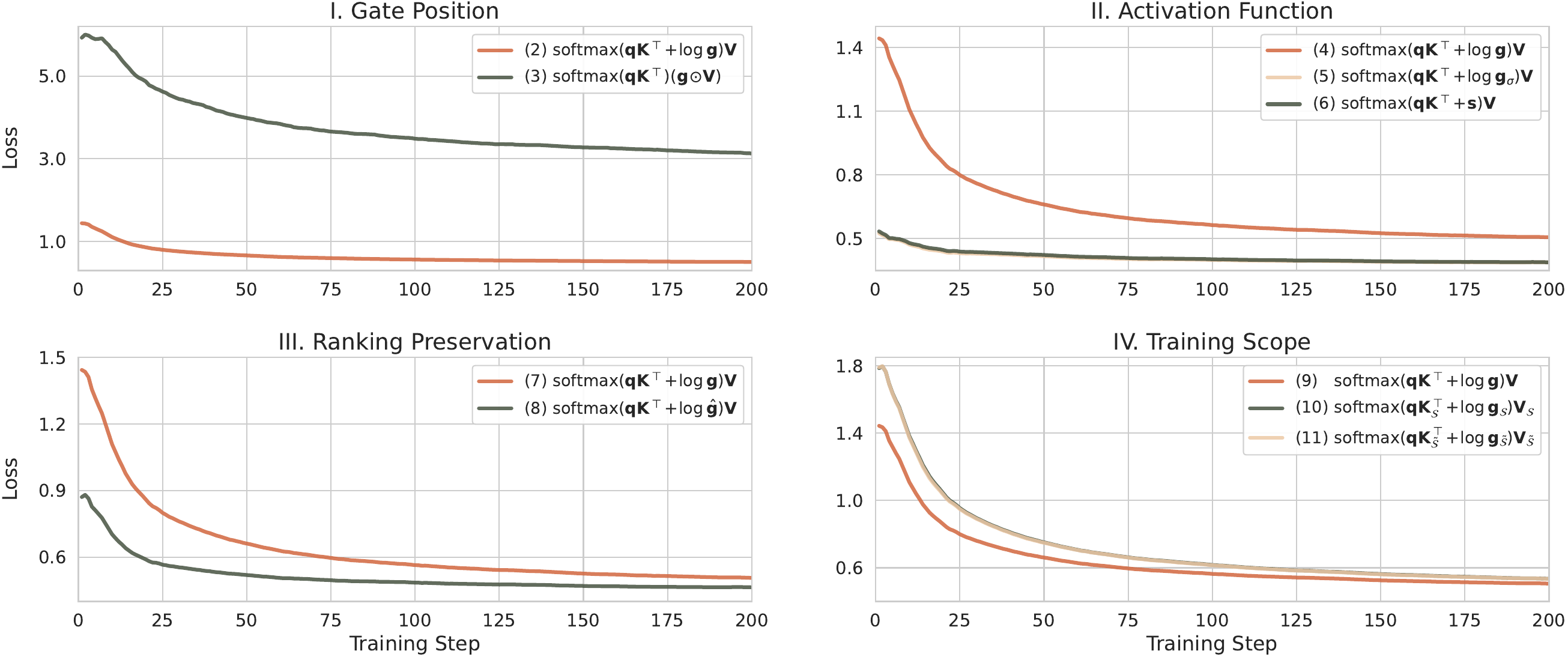}
        \caption{}
    \end{subfigure}

    \begin{subfigure}{0.99\linewidth}
        \centering
        \includegraphics[width=\linewidth]{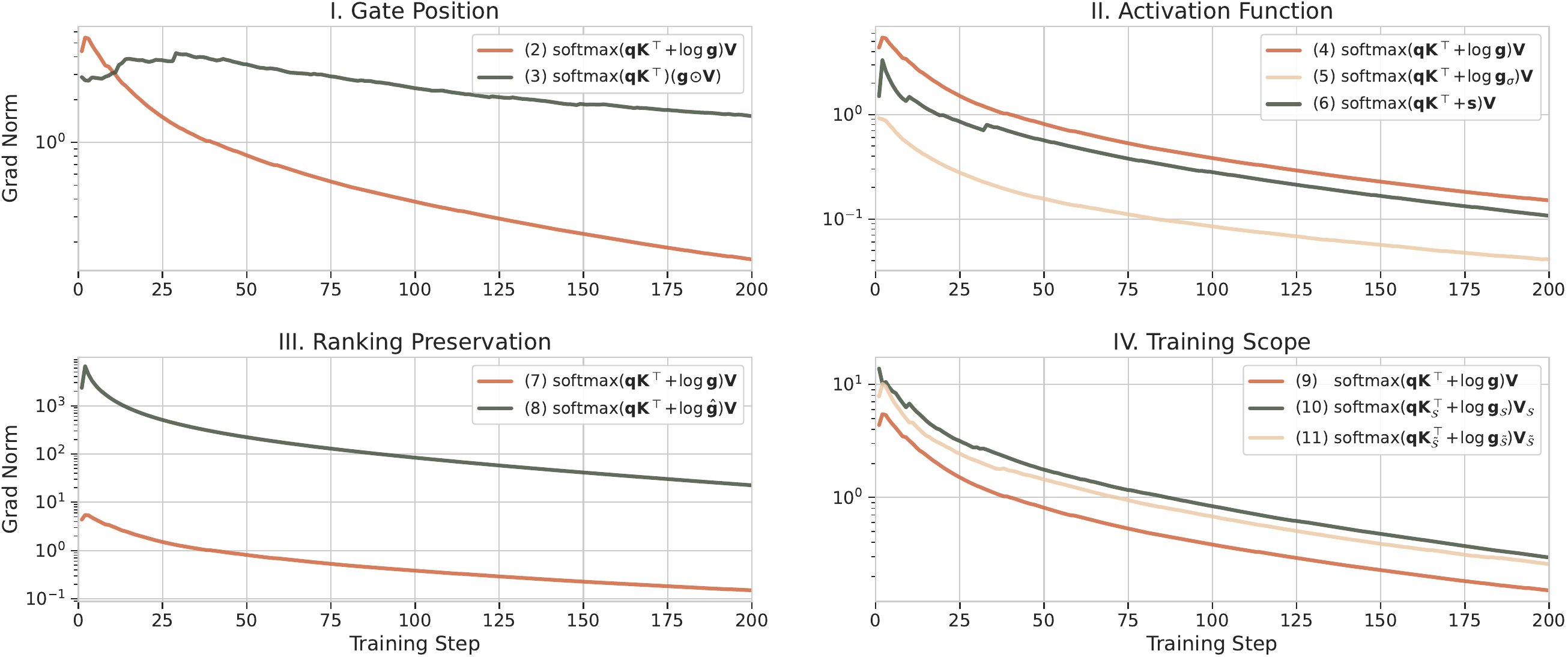}
        \caption{}
    \end{subfigure}

    \caption{Training dynamics for different gated sparsification designs.
    Inner $\operatorname{softmax}$ gate yields lower training loss than outer gate.
    Independent $\operatorname{sigmoid}$ gates and raw-logit injection achieve lower loss than normalized $\operatorname{softmax}$ gates.
    Hard gating reduces the training loss, but substantially increases the gradient norm.
    Sparse scope training starts with a higher loss but gradually approaches full scope training, while introducing noise yields little difference.}
    \label{fig:loss_ab}
\end{figure}

We provide the loss and gradient norm curves for the controlled ablation studies in Section~\ref{subse:ranking_learn} in \Cref{fig:loss_ab}. The training dynamics reveal distinct optimization behaviors across designs. The inner-softmax gate consistently achieves lower training loss than the outer-gate formulation. Independent sigmoid gates and raw-logit injection achieve even lower losses, but fail to improve sparsification, as both effectively approach ungated attention and learn little discriminative block ranking. Similarly, although the STE formulation yields lower training loss, its final sparsification performance remains inferior, indicating that training loss alone is insufficient to assess ranking quality. Hard gating further reduces loss while substantially increasing the gradient norm, suggesting more aggressive optimization. Sparse-scope training starts with higher loss but gradually approaches full-scope training, indicating that the two scopes can eventually reach comparable solutions despite different gradient dynamics. Overall, effective selector training requires not only loss minimization but also preservation of a meaningful and discriminative block ranking signal.

\section{Limitation}
\label{app:limitation}

We further explore the scalability of the learned selector to RULER \cite{hsieh2024ruler}. Specifically, we sample 0.5B tokens from the \texttt{allenai/dolma3\_longmino\_mix-100B-1125} dataset~\footnote{\url{https://huggingface.co/datasets/allenai/dolma3_longmino_mix-100B-1125}} and train the selector on 64K sequences using a packing strategy. During evaluation, we use the non-thinking mode and extend the context window to 128K with YaRN. As shown in \Cref{tab:ruler}, while \our consistently improves over SeerAttention-R under shorter context lengths, its performance still degrades substantially as the context length increases, leaving a considerable gap from full attention. We believe this limitation primarily stems from the use of pooling based block summaries, which may struggle to preserve fine grained, localized information, such as needle-like signals, within each block, especially as the context length increases. 
Designing more expressive yet efficient selectors that can capture fine-grained information under context block compression is therefore an important direction for future work.

\begin{table}[h]
    \centering
    \caption{
        Evaluation results on the RULER benchmark.
    }
    \label{tab:ruler}
    \resizebox{\textwidth}{!}{
    \begin{tabular}{ll
        rrrrrr
        rrrrrr
        rrrrrr}
    \toprule
    \multirow{2}{*}{\textbf{Budget}} &
    \multirow{2}{*}{\textbf{Method}} &
    \multicolumn{6}{c}{\textbf{Qwen3-4B}} &
    \multicolumn{6}{c}{\textbf{Qwen3-8B}} &
    \multicolumn{6}{c}{\textbf{Qwen3-14B}}
    \\
    \cmidrule(lr){3-8}
    \cmidrule(lr){9-14}
    \cmidrule(lr){15-20}
    &
    &
    \textbf{4K} & \textbf{8K} & \textbf{16K} &
    \textbf{32K} & \textbf{64K} & \textbf{128K}
    &
    \textbf{4K} & \textbf{8K} & \textbf{16K} &
    \textbf{32K} & \textbf{64K} & \textbf{128K}
    &
    \textbf{4K} & \textbf{8K} & \textbf{16K} &
    \textbf{32K} & \textbf{64K} & \textbf{128K}
    \\
    \midrule

    \textit{Full}
    & Full Attn
    & 92.95 & 91.27 & 86.61 & 79.25 & 73.17 & 63.81
    & 93.10 & 90.13 & 89.59 & 87.04 & 77.25 & 73.87
    & 95.28 & 92.71 & 92.91 & 92.15 & 92.19 & 82.23
    \\

    \midrule

    \multirow{2}{*}{2048}
    & SeerAttn-R
    & 92.08 & 79.28 & 56.01 & 30.29 & 18.12 & 13.09
    & 92.15 & 80.16 & 63.76 & 42.16 & 24.49 & 13.99
    & 94.40 & 84.50 & 62.80 & 44.73 & 30.73 & 14.95
    \\
    
    & \our
    & {92.82} & 78.22 & 63.94 & 39.54 & 25.65 & 17.92
    & {92.45} & 81.39 & 69.34 & 46.94 & 30.31 & 19.38
    & 94.43 & 85.10 & 69.01 & 54.76 & 34.12 & 23.80
    \\

    \midrule

    \multirow{2}{*}{4096}
    & SeerAttn-R
    & 93.49 & 87.90 & 72.93 & 47.41 & 27.96 & 16.29
    & 92.96 & 90.13 & 77.39 & 57.92 & 37.43 & 18.75
    & 95.28 & 91.62 & 76.77 & 64.96 & 42.34 & 22.36
    \\

    & \our
    & {93.49} & 89.09 & 75.72 & 51.90 & 35.53 & 21.87
    & {92.96} & 87.96 & 79.88 & 63.50 & 39.18 & 23.46
    & {95.28} & 91.52 & 83.35 & 66.82 & 45.92 & 29.95
    \\

    \bottomrule
    \end{tabular}
    }
\end{table}

\section{Final Performance of Sparse vs.\ Full Training Scope}
\label{app:scope_final}
The gradient analysis in Appendix~\ref{app:deriv_routing_scope} shows that sparse scope updates unselected blocks only indirectly, which slows early optimization but does not change the final ranking quality.
\Cref{tab:scope_final} reports the final accuracy of full scope and sparse scope across Qwen3-4B/8B/14B and token budgets of 1024, 2048, and 4096.
Consistent with the training dynamics in \Cref{tab:ablation_main}, sparse scope matches full scope at convergence across all model scales and budgets, while incurring lower training cost.

\begin{table}[h]
\centering
\caption{Final accuracy (\%) of full training scope versus sparse training scope across model scales and token budgets. Standard deviations are shown in parentheses.}
\label{tab:scope_final}
\resizebox{\textwidth}{!}{
\begin{tabular}{ll lll lll lll lll}
\toprule
\multirow{2}{*}{\textbf{Budget}} & \multirow{2}{*}{\textbf{Train scope}} &
\multicolumn{3}{c}{\textbf{MATH500}} &
\multicolumn{3}{c}{\textbf{GPQA-Diamond}} &
\multicolumn{3}{c}{\textbf{AIME24}} &
\multicolumn{3}{c}{\textbf{AIME25}} \\
\cmidrule(lr){3-5} \cmidrule(lr){6-8} \cmidrule(lr){9-11} \cmidrule(lr){12-14}
& & \textbf{4B} & \textbf{8B} & \textbf{14B}
  & \textbf{4B} & \textbf{8B} & \textbf{14B}
  & \textbf{4B} & \textbf{8B} & \textbf{14B}
  & \textbf{4B} & \textbf{8B} & \textbf{14B} \\
\midrule
\multirow{2}{*}{1024}
& full
& \valstd{91.33}{1.1} & \valstd{91.71}{1.1} & \valstd{93.08}{1.0}
& \valstd{51.42}{2.9} & \valstd{55.62}{2.8} & \valstd{61.77}{2.9}
& - & - & -
& - & - & - \\
& sparse
& \valstd{90.65}{1.1} & \valstd{91.27}{1.1} & \valstd{92.93}{1.0}
& \valstd{50.41}{2.8} & \valstd{53.17}{2.9} & \valstd{61.14}{2.9}
& - & - & -
& - & - & - \\
\midrule
\multirow{2}{*}{2048}
& full
& \valstd{93.00}{1.0} & \valstd{93.09}{1.1} & \valstd{93.56}{1.0}
& \valstd{54.40}{2.9} & \valstd{58.43}{2.9} & \valstd{64.14}{2.8}
& \valstd{68.13}{6.7} & \valstd{70.73}{6.6} & \valstd{76.67}{6.7}
& \valstd{55.39}{7.3} & \valstd{59.27}{7.3} & \valstd{64.04}{7.2} \\
& sparse
& \valstd{93.47}{1.0} & \valstd{93.17}{1.0} & \valstd{93.54}{1.0}
& \valstd{54.86}{2.8} & \valstd{58.74}{2.8} & \valstd{65.09}{2.8}
& \valstd{68.85}{6.6} & \valstd{70.73}{6.7} & \valstd{77.76}{6.6}
& \valstd{56.38}{7.3} & \valstd{58.41}{7.2} & \valstd{64.19}{7.1} \\
\midrule
\multirow{2}{*}{4096}
& full
& \valstd{93.10}{1.0} & \valstd{93.40}{1.0} & \valstd{94.12}{1.0}
& \valstd{54.80}{2.8} & \valstd{60.32}{2.8} & \valstd{64.39}{2.9}
& \valstd{71.17}{6.6} & \valstd{73.52}{6.7} & \valstd{78.49}{6.6}
& \valstd{61.30}{7.3} & \valstd{64.24}{7.1} & \valstd{67.58}{7.0} \\
& sparse
& \valstd{93.67}{1.0} & \valstd{94.23}{1.0} & \valstd{95.23}{0.8}
& \valstd{55.11}{2.9} & \valstd{60.42}{2.8} & \valstd{65.03}{2.8}
& \valstd{71.72}{6.8} & \valstd{73.49}{6.7} & \valstd{78.28}{6.4}
& \valstd{59.97}{7.3} & \valstd{63.41}{7.2} & \valstd{67.29}{6.9} \\
\bottomrule
\end{tabular}
}
\end{table}

\section{Gradient Derivation}
\label{app:derivation}

\subsection{Gradient Derivation for Gate Position}
\label{app:deriv_gate_position}
We derive the gate gradients for outer and inner gate injection.
First, we let
\[
    z_i = \mathbf{q}\mathbf{k}_i^\top,
    \qquad
    p_i = \frac{\exp(z_i)}{\sum_j \exp(z_j)},
    \qquad
    b(i) = m \;\; \text{if } i\in B_m .
\]

\paragraph{Outer gate.}
For outer gating, the attention probability is computed before applying the gate:
\[
    \mathbf{o}_{outer}
    =
    \sum_i p_i g_{b(i)}\mathbf{v}_i .
\]
Since $p_i$ does not depend on $g_m$, the outer-gate gradient is
\[
\begin{aligned}
    dg_m^{outer}
    &=
    \frac{\partial \mathcal{L}}{\partial g_m}
    =
    d\mathbf{o}^{\top}
    \frac{\partial \mathbf{o}_{outer}}{\partial g_m}
    \\
    &=
    d\mathbf{o}^{\top}
    \sum_i
    p_i
    \frac{\partial g_{b(i)}}{\partial g_m}
    \mathbf{v}_i
    \\
    &=
    \sum_{i\in B_m}
    p_i\,
    d\mathbf{o}^{\top}\mathbf{v}_i .
\end{aligned}
\]

\paragraph{Inner gate.}
For inner gating, the gate is injected into the attention logits:
\[
    \mathbf{o}_{inner}
    =
    \sum_i \tilde p_i \mathbf{v}_i,
    \qquad
    \tilde p_i
    =
    \frac{g_{b(i)}\exp(z_i)}
    {\sum_j g_{b(j)}\exp(z_j)} .
\]

For a block gate $g_m$, its derivative is
\[
\begin{aligned}
    \frac{\partial \tilde p_i}{\partial g_m}
    &=
    \frac{
    \mathbb{I}[b(i)=m]\exp(z_i)
    \sum_j g_{b(j)}\exp(z_j)
    -
    g_{b(i)}\exp(z_i)
    \sum_{j\in B_m}\exp(z_j)
    }
    {\left(\sum_j g_{b(j)}\exp(z_j)\right)^2}
    \\
    &=
    \frac{
    \mathbb{I}[b(i)=m]\exp(z_i)
    }
    {\sum_j g_{b(j)}\exp(z_j)}
    -
    \frac{
    g_{b(i)}\exp(z_i)
    }
    {\sum_j g_{b(j)}\exp(z_j)}
    \cdot
    \frac{
    \sum_{j\in B_m}\exp(z_j)
    }
    {\sum_j g_{b(j)}\exp(z_j)}
    \\
    &=
    \frac{\mathbb{I}[b(i)=m]}{g_m}
    \frac{
    g_{b(i)}\exp(z_i)
    }
    {\sum_j g_{b(j)}\exp(z_j)}
    -
    \frac{
    g_{b(i)}\exp(z_i)
    }
    {\sum_j g_{b(j)}\exp(z_j)}
    \cdot
    \sum_{j\in B_m}
    \frac{\exp(z_j)}
    {\sum_r g_{b(r)}\exp(z_r)}
    \\
    &=
    \frac{\mathbb{I}[b(i)=m]}{g_m}\tilde p_i
    -
    \tilde p_i
    \sum_{j\in B_m}
    \frac{1}{g_m}
    \frac{
    g_m\exp(z_j)
    }
    {\sum_r g_{b(r)}\exp(z_r)}
    \\
    &=
    \frac{\mathbb{I}[b(i)=m]}{g_m}\tilde p_i
    -
    \tilde p_i
    \sum_{j\in B_m}
    \frac{\tilde p_j}{g_m}
    \\
    &=
    \tilde p_i
    \left(
        \frac{\mathbb{I}[b(i)=m]}{g_m}
        -
        \sum_{j\in B_m}\frac{\tilde p_j}{g_m}
    \right).
\end{aligned}
\]

Therefore,
\[
\begin{aligned}
    dg_m^{inner}
    &=
    \frac{\partial \mathcal{L}}{\partial g_m}
    =
    d\mathbf{o}^{\top}
    \frac{\partial \mathbf{o}_{inner}}{\partial g_m}
    \\
    &=
    d\mathbf{o}^{\top}
    \sum_i
    \frac{\partial \tilde p_i}{\partial g_m}
    \mathbf{v}_i
    \\
    &=
    d\mathbf{o}^{\top}
    \sum_i
    \tilde p_i
    \left(
        \frac{\mathbb{I}[b(i)=m]}{g_m}
        -
        \sum_{j\in B_m}\frac{\tilde p_j}{g_m}
    \right)
    \mathbf{v}_i
    \\
    &=
    d\mathbf{o}^{\top}
    \left[
    \frac{1}{g_m}
    \sum_{i\in B_m}
    \tilde p_i \mathbf{v}_i
    -
    \frac{1}{g_m}
    \left(
        \sum_{j\in B_m}\tilde p_j
    \right)
    \sum_i \tilde p_i\mathbf{v}_i
    \right]
    \\
    &=
    \frac{1}{g_m}
    d\mathbf{o}^{\top}
    \sum_{i\in B_m}
    \tilde p_i
    \left(
        \mathbf{v}_i-\mathbf{o}_{inner}
    \right)
    \\
    &=
    \sum_{i\in B_m}
    \frac{\tilde p_i}{g_m}
    d\mathbf{o}^{\top}
    \left(
        \mathbf{v}_i-\mathbf{o}_{inner}
    \right).
\end{aligned}
\]

\subsection{Gradient Derivation for Ranking Preservation}
\label{app:deriv_ranking_pres}
Both variants use inner gate injection, so both inherit the inner-gate gradient from Appendix~\ref{app:deriv_gate_position}, with $z_i=\mathbf{q}\mathbf{k}_i^\top$.
They differ only in the attention weight that the forward normalizer assigns to each token.

\paragraph{Soft gating.}
All $C$ blocks stay active, so the normalizer runs over the full context and block $B_m$ appears in its own normalizer:
\[
    \tilde p_i^{soft}
    =
    \frac{g_{b(i)}\exp(z_i)}
    {\sum_j g_{b(j)}\exp(z_j)}
    \le 1,
    \qquad
    dg_m
    =
    \sum_{i\in B_m}
    \frac{\tilde p_i^{soft}}{g_m}\,
    d\mathbf{o}^{\top}
    \left(
        \mathbf{v}_i-\mathbf{o}
    \right).
\]
Hence $\tilde p_i^{soft}\le 1$ for every token, and the gate gradient $dg_m$ stays bounded.

\paragraph{Hard gating with STE.}
The unselected blocks satisfy $\log g^{hard}_m=-\infty$, so the forward normalizer sums only over the selected set $\mathcal{S}=\bigcup_{m\in\mathcal{I}}B_m$.
STE still propagates a gate gradient to a non-selected block $m\notin\mathcal{I}$, whose ranking signal evaluates the attention mass $\tilde p_i^{hard}$ its tokens would receive.
Because $B_m$ is excluded from $\operatorname{LSE}_{\mathcal{S}}$, this weight has no upper bound:
\[
    \tilde p_i^{hard}
    =
    \frac{\exp(z_i)}
    {\sum_{j\in\mathcal{S}}\exp(z_j)}
    =
    \exp\!\left(z_i-\operatorname{LSE}_{\mathcal{S}}\right)
    \le
    \exp\!\left(z_i-\max_{j\in\mathcal{S}}z_j\right),
    \qquad
    \operatorname{LSE}_{\mathcal{S}}
    =
    \log\sum_{j\in\mathcal{S}}\exp(z_j),
\]
which grows exponentially once a dropped block outscores the selected set, $z_i>\max_{j\in\mathcal{S}}z_j$.
The unbounded $\tilde p_i^{hard}$ enters $dg_m$ and, accumulated over all dropped blocks, tokens, and heads, inflates the gate gradient.

\subsection{Gradient Derivation for Training Scope}
\label{app:deriv_routing_scope}
We derive the score gradients under sparse and full scopes.
Let the activation function $\operatorname{softmax}$.

For any gate gradient $dg_\ell$, we first give the $\operatorname{softmax}$ backward formulation
\[
\begin{aligned}
    ds_m
    &=
    \frac{\partial \mathcal{L}}{\partial s_m}
    =
    \sum_{\ell=1}^{C}
    \frac{\partial \mathcal{L}}{\partial g_\ell}
    \frac{\partial g_\ell}{\partial s_m}
    \\
    &=
    \sum_{\ell=1}^{C}
    dg_\ell\,
    g_\ell
    \left(
        \mathbb{I}[\ell=m]-g_m
    \right)
    \\
    &=
    g_m dg_m
    -
    g_m
    \sum_{\ell=1}^{C} g_\ell dg_\ell
    \\
    &=
    g_m
    \left(
        dg_m
        -
        \sum_{\ell=1}^{C} g_\ell dg_\ell
    \right).
\end{aligned}
\]

\paragraph{Sparse routing scope.}
Let $\mathcal{I}=\text{Top-}K(\mathbf{s},K)$ denote the selected block indices.
Under sparse routing scope, only selected blocks participate in the gated attention computation.
For an unselected block $m\notin\mathcal{I}$, the output does not depend on $g_m$, so $dg_m^{sparse}=0$.

Moreover, only selected blocks have nonzero gate gradients:
\[
    dg_\ell^{sparse}=0,
    \qquad
    \ell\notin\mathcal{I}.
\]
Therefore, for $m\notin\mathcal{I}$,
\[
\begin{aligned}
    ds_m^{sparse}
    &=
    g_m
    \left(
        dg_m^{sparse}
        -
        \sum_{\ell=1}^{C}
        g_\ell dg_\ell^{sparse}
    \right)
    \\
    &=
    -g_m
    \sum_{\ell=1}^{C}
    g_\ell dg_\ell^{sparse}
    \\
    &=
    -g_m
    \sum_{\ell\in\mathcal{I}}
    g_\ell dg_\ell^{sparse}.
\end{aligned}
\]

\paragraph{Full routing scope.}
Under full routing scope, all blocks participate in the gated attention computation during training.
Therefore, an unselected block $m\notin\mathcal{I}$ still has its own gate gradient $dg_m^{full}$.
Using the same softmax backward identity,
\[
\begin{aligned}
    ds_m^{full}
    &=
    g_m
    \left(
        dg_m^{full}
        -
        \sum_{\ell=1}^{C}
        g_\ell dg_\ell^{full}
    \right)
    \\
    &=
    -g_m
    \left(
        -dg_m^{full}
        +
        \sum_{\ell=1}^{C}
        g_\ell dg_\ell^{full}
    \right).
\end{aligned}
\]

\definecolor{codeblue}{HTML}{8A8A8A}
\definecolor{codekw}{HTML}{CB7C5E}
\definecolor{codesign}{HTML}{A5432D}
\definecolor{codefunc}{HTML}{D1917F}

\lstdefinelanguage{PythonFuncColor}{
  language=Python,
  keywordstyle=\color{codesign}\bfseries,
  commentstyle=\color{codeblue},  %
  stringstyle=\color{orange},
  showstringspaces=false,
  basicstyle=\ttfamily\small,
  literate=
    {*}{{\color{codesign}* }}{1}
    {dataloader}{{\color{codefunc}dataloader}}{1}
    {sample_t_r}{{\color{codefunc}sample\_t\_r}}{1}
    {randn}{{\color{codefunc}randn}}{1}
    {randn_like}{{\color{codefunc}randn\_like}}{1}
    {jvp}{{\color{codefunc}jvp}}{1}
    {stopgrad}{{\color{codefunc}stopgrad}}{1}
    {metric}{{\color{codefunc}metric}}{1}
}

\lstset{
  language=PythonFuncColor,
  backgroundcolor=\color{white},
  basicstyle=\fontsize{9pt}{9.9pt}\ttfamily\selectfont,
  columns=fullflexible,
  breaklines=true,
  captionpos=b,
  frame=none
}

\section{Kernel Implementation}
\label{sec:app_kernel}

We provide a custom Triton implementation for the sparse scope soft gated attention used during SAS training.
The block gate is organized as a table $\log\mathbf{G}$, where each query position and attention head holds one log gate value for every previous context block.
Block selection is encoded by a per query threshold $\tau$: a block is \emph{selected} when its log gate reaches the threshold ($\log g_m\ge\tau$), which realizes the Top-$K$ set $\mathcal{I}$ without an explicit sort. Non selected blocks are masked out of the softmax so that only the selected set $\mathcal{S}$ participates.

For simplicity of presentation, the pseudocode omits the grouped query attention, and variable length packing that the full implementation supports, and focuses on the gate injection and the sparse scope.

\paragraph{Forward pass.}
The kernel traverses causal key value blocks in a tiled manner.
The current (self) block always uses a unit gate and a causal mask, so local causal attention is always preserved.
For each earlier block, the kernel maps key positions to their block index, loads the corresponding log gate, and adds it to the $\mathbf{q}\mathbf{K}_{\mathcal{S}}^\top$ scores before the online softmax update; blocks below the threshold are masked to $-\infty$ and an entire block is skipped when no query in the tile selects it.

\paragraph{Backward pass.}
We follow the standard recomputation based attention backward procedure.
The kernel first computes the row-wise correction term $\mathbf{\Delta}=\sum_d\mathbf{O}_d\cdot d\mathbf{O}_d$.
It then recomputes the gated attention probabilities from the saved log-sum-exp values and uses them to obtain gradients for $\mathbf{Q}$, $\mathbf{K}$, and $\mathbf{V}$.
In addition, the gradient of each block log gate is accumulated by summing the score gradients over all tokens inside the corresponding selected block, producing a compact block-level gradient tensor that is propagated back to the selector.
As a result, the implementation supports sparse scope gated attention training without storing the dense attention matrix.

\begin{algorithm}
\caption{Forward Pass of Fused SAS Kernel}
\label{alg:soft_gate_fwd}
\begin{lstlisting}[language=python]
def gated_scores(q, k, logG, tau, q_tile, kv_tile, B, sm_scale, self_block):
    S = sm_scale * matmul(q, k.T)
    b = block_id(kv_tile, B)
    if self_block:
        # current block: unit gate (log 1 = 0), causal mask
        return apply_causal_mask(S, q_tile, kv_tile)
    log_g = load(logG, q_tile, b)               # [q_rows] per-query block gate
    active = log_g >= load(tau, q_tile)         # selected blocks only (Top-K set)
    S = S + broadcast(log_g, kv_tile)           # inject gate inside softmax
    return where(active, S, -inf)               # mask out non-selected blocks

def soft_gate_fwd(Q, K, V, logG, tau, B, sm_scale):
    O, LSE = empty_output(Q, V), empty_lse(Q)

    for q_tile in tiles(Q):
        q = load(Q, q_tile)
        m, l = full(q.rows, -inf), zeros(q.rows)
        acc = zeros(q.rows, V.dim)

        # Phase 1: self block (causal) ; Phase 2: earlier blocks (sparse)
        for kv_tile in causal_tiles(K, q_tile):
            self_block = in_self_block(kv_tile, q_tile, B)
            if (not self_block) and no_query_selects(logG, tau, q_tile, kv_tile, B):
                continue                        # skip fully non-selected block
            k, v = load(K, kv_tile), load(V, kv_tile)
            S = gated_scores(q, k, logG, tau, q_tile, kv_tile, B, sm_scale, self_block)

            m_new = maximum(m, rowmax(S))
            P = exp(S - m_new[:, None])
            acc = acc * exp(m - m_new)[:, None] + matmul(P, v)
            l = l * exp(m - m_new) + rowsum(P)
            m = m_new

        O[q_tile] = acc / l[:, None]
        LSE[q_tile] = m + log(l)

    return O, LSE
\end{lstlisting}
\end{algorithm}

\begin{algorithm}
\caption{Backward Pass of Fused SAS Kernel}
\label{alg:soft_gate_bwd}
\begin{lstlisting}[language=python]
def soft_gate_bwd(Q, K, V, logG, tau, O, dO, LSE, B, sm_scale):
    dQ, dK, dV = zeros_like(Q), zeros_like(K), zeros_like(V)
    dLogG, Delta = zeros_like(logG), rowsum(O * dO)

    # ---- Pass 1: dQ and dLogG (loop q outer, kv inner) ----
    for q_tile in tiles(Q):
        q, do = load(Q, q_tile), load(dO, q_tile)
        lse, delta = load(LSE, q_tile), load(Delta, q_tile)
        dq = zeros_like(q)

        for kv_tile in causal_tiles(K, q_tile):
            self_block = in_self_block(kv_tile, q_tile, B)
            if (not self_block) and no_query_selects(logG, tau, q_tile, kv_tile, B):
                continue
            k, v = load(K, kv_tile), load(V, kv_tile)
            S = gated_scores(q, k, logG, tau, q_tile, kv_tile, B, sm_scale, self_block)

            P = exp(S - lse[:, None])
            dS = P * (matmul(do, v.T) - delta[:, None])
            dq += sm_scale * matmul(dS, k)
            if not self_block:
                for b in blocks(kv_tile, B):
                    # gate gradient: sum dS over kv tokens in the selected block
                    dLogG[q_tile, b] += rowsum(dS[:, tokens(kv_tile, b)])

        dQ[q_tile] = dq

    # ---- Pass 2: dK and dV (loop kv outer, q inner) ----
    for kv_tile in tiles(K):
        k, v = load(K, kv_tile), load(V, kv_tile)
        dk, dv = zeros_like(k), zeros_like(v)

        for q_tile in attending_q_tiles(Q, kv_tile):
            self_block = in_self_block(kv_tile, q_tile, B)
            if (not self_block) and no_query_selects(logG, tau, q_tile, kv_tile, B):
                continue
            q, do = load(Q, q_tile), load(dO, q_tile)
            lse, delta = load(LSE, q_tile), load(Delta, q_tile)
            S = gated_scores(q, k, logG, tau, q_tile, kv_tile, B, sm_scale, self_block)

            P = exp(S - lse[:, None])
            dS = P * (matmul(do, v.T) - delta[:, None])
            dk += sm_scale * matmul(dS.T, q)
            dv += matmul(P.T, do)

        dK[kv_tile], dV[kv_tile] = dk, dv

    return dQ, dK, dV, dLogG
\end{lstlisting}
\end{algorithm}

\end{document}